\documentclass[runningheads]{llncs}
\usepackage[dvipsnames, table]{xcolor}
\usepackage{graphicx}
\usepackage{tikz}
\usepackage{comment}
\usepackage{amsmath,amssymb} % define this before the line numbering.

\usepackage{subcaption}
\usepackage{graphicx}
\usepackage{amsmath}
\usepackage{amssymb}
\usepackage{mathtools}
\usepackage{booktabs}
\usepackage{kotex}
\usepackage{fontawesome5}
\usepackage{algorithm}
\usepackage{algpseudocode}
\usepackage{pythonhighlight}
\lstnewenvironment{MyPython}[1][]{\lstset{style=mypython, frame=none, #1}}{}
\algrenewcommand\algorithmiccomment[1]{\textcolor{RoyalBlue}{ \textbf{\footnotesize{\textcolor{RoyalBlue}{\\\# #1}}}}}

\newcommand{\NOTATION}[1]{\textcolor{RoyalBlue}{\tt\textit{\textbf{\footnotesize{\textcolor{Maroon}{\\\# #1}}}}}}

\newcommand{\CODE}[1]{\tt{\small{\\#1}}}

\makeatletter
\newenvironment{Configs}[1][H]{%
    \renewcommand{\ALG@name}{Configs}
   \begin{algorithm}[#1]%
  }{\end{algorithm}}
\makeatother

\makeatletter
\newenvironment{pseudocode}[1][H]{%
    \renewcommand{\ALG@name}{Pseudo-code}
   \begin{algorithm}[#1]%
  }{\end{algorithm}}
\makeatother

\usepackage[pagebackref,breaklinks,colorlinks=true,linkcolor=RoyalBlue,citecolor=RoyalBlue]{hyperref}
\definecolor{myblue}{HTML}{d4e6f1}
\definecolor{orcidlogocol}{rgb}{0.65, 0.807, 0.223}

\newcommand{\orcid}[1]{$\,$\href{https://orcid.org/#1}{\textcolor{orcidlogocol}{\faOrcid}}}

\usepackage{ctable} % for \specialrule command

\usepackage[accsupp]{axessibility}  % Improves PDF readability for those with disabilities.

\begin{document}
% \renewcommand\thelinenumber{\color[rgb]{0.2,0.5,0.8}\normalfont\sffamily\scriptsize\arabic{linenumber}\color[rgb]{0,0,0}}
% \renewcommand\makeLineNumber {\hss\thelinenumber\ \hspace{6mm} \rlap{\hskip\textwidth\ \hspace{6.5mm}\thelinenumber}}
% \linenumbers
\pagestyle{headings}
\mainmatter
\def\ECCVSubNumber{5082}  % Insert your submission number here

\title{Contrastive Vicinal Space for Unsupervised Domain Adaptation} % Replace with your title

% INITIAL SUBMISSION 
\begin{comment}
\titlerunning{ECCV-22 submission ID \ECCVSubNumber} 
\authorrunning{ECCV-22 submission ID \ECCVSubNumber} 
\author{Anonymous ECCV submission}
\institute{Paper ID \ECCVSubNumber}
\end{comment}
%******************

% CAMERA READY SUBMISSION
% \begin{comment}
% \titlerunning{Abbreviated paper title}
% If the paper title is too long for the running head, you can set
% an abbreviated paper title here
%
\author{Jaemin Na\inst{1}\orcid{0000-0002-8604-2839} \and
Dongyoon Han\inst{2}\orcid{0000-0002-9130-8195} \and
Hyung Jin Chang\inst{3}\orcid{0000-0001-7495-9677} \and
Wonjun Hwang\inst{1}\orcid{0000-0001-8895-0411}
% $^1$Ajou University, Korea, $^2$Kyungpook National University, Korea, $^3$University of Birmingham, UK
}
\authorrunning{J. Na et al.}

% \institute{Ajou University, Korea \and NAVER AI Lab \and University of Birmingham, UK
% \email{lncs@springer.com}\\
% \url{http://www.springer.com/gp/computer-science/lncs} \and
% ABC Institute, Rupert-Karls-University Heidelberg, Heidelberg, Germany\\
% \email{\{abc,lncs\}@uni-heidelberg.de}}

\institute{
$^1$Ajou University, Korea \quad
% \email{\{osial46,wjhwang\}@ajou.ac.kr}\\
% \and
$^2$NAVER AI Lab \quad
% \email{dongyoon.han@navercorp.com}\\
% \and
$^3$University of Birmingham, UK
\email{osial46@ajou.ac.kr}, \email{dongyoon.han@navercorp.com},\\ \email{h.j.chang@bham.ac.uk}, \email{wjhwang@ajou.ac.kr}\\
% \email{\{abc,lncs\}@uni-heidelberg.de}
}
% \end{comment}
%******************
\maketitle

% \vspace{-3mm}
\begin{abstract}
% Utilizing vicinal space between the source and target domains is one of the recent unsupervised domain adaptation approaches.
Recent unsupervised domain adaptation methods have utilized vicinal space between the source and target domains. However, the equilibrium collapse of labels, a problem where the source labels are dominant over the target labels in the predictions of vicinal instances, has never been addressed. In this paper, we propose an instance-wise minimax strategy that minimizes the entropy of high uncertainty instances in the vicinal space to tackle the stated problem. We divide the vicinal space into two subspaces through the solution of the minimax problem: {contrastive space} and {consensus space}. In the {contrastive space}, inter-domain discrepancy is mitigated by constraining instances to have contrastive views and labels, and the {consensus space} reduces the confusion between intra-domain categories. The effectiveness of our method is demonstrated on public benchmarks, including Office-31, Office-Home, and VisDA-C, achieving state-of-the-art performances. We further show that our method outperforms the current state-of-the-art methods on PACS, which indicates that our instance-wise approach works well for multi-source domain adaptation as well. Code is available at \url{https://github.com/NaJaeMin92/CoVi}.

\keywords{Unsupervised Domain Adaptation, Equilibrium Collapse, Contrastive Vicinal Space}
\end{abstract}
\vspace{-5mm}

\begin{figure}[t]
\centering
% \vspace{3mm}
\includegraphics[width=0.7\columnwidth]{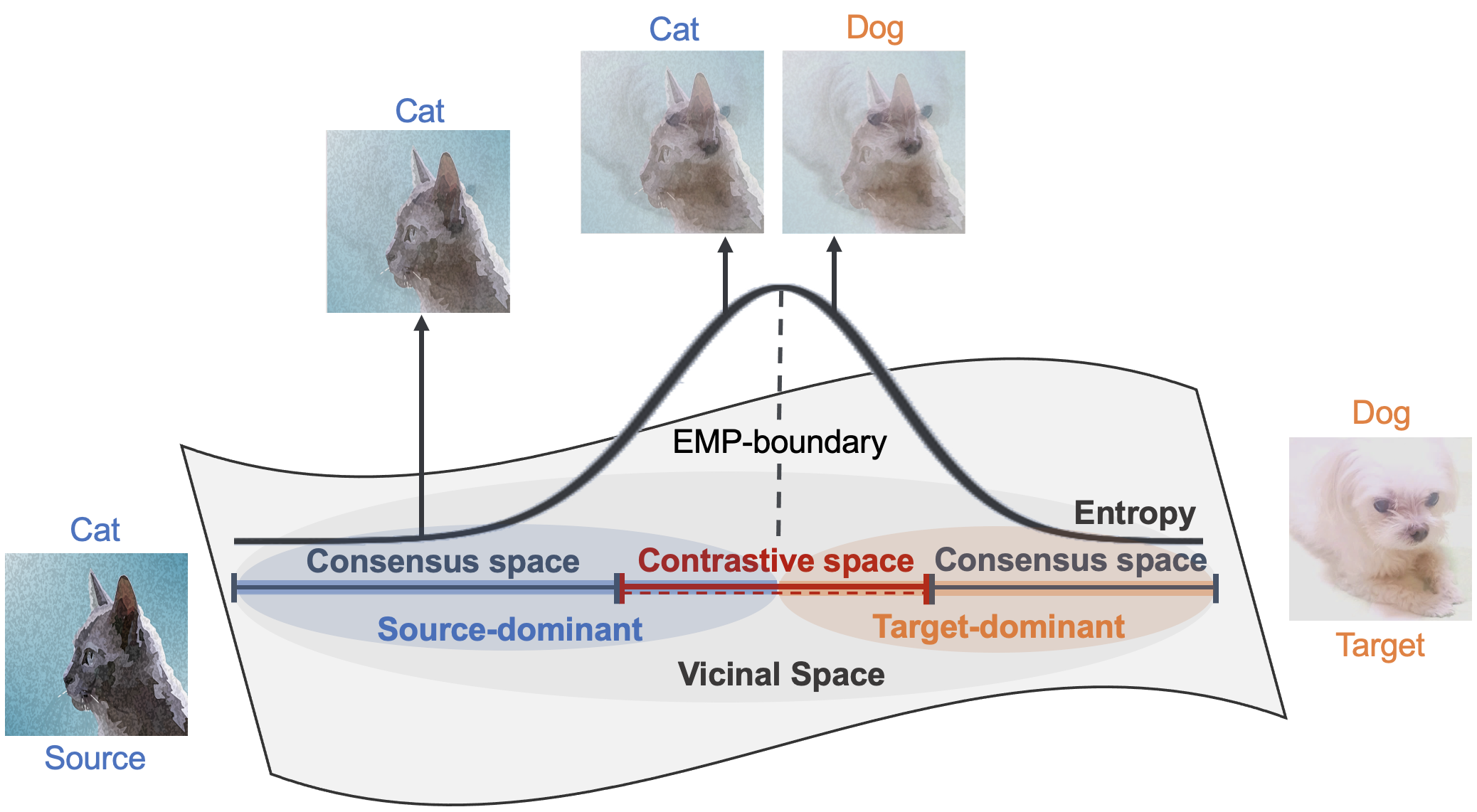}
\vspace{-3mm}
\caption{\textbf{Overview.} Vicinal space between the source and target domains is divided into contrastive space and consensus space. Our methodology alleviates inter-domain discrepancy in the contrastive space and simultaneously resolves intra-domain categorical confusion in the consensus space.
}
\label{fig:01}
\vspace{-6mm}
\end{figure}
% \vspace{-4mm}

\section{Introduction}

Unsupervised domain adaptation (UDA) aims to adapt a model trained on a labeled source domain to an unlabeled target domain. One of the most important problems to solve in UDA is the domain shift~\cite{shimodaira2000improving} (\textit{i.e.,} distribution shift) problem. The domain shift arises from the change in the data distribution between the training domain (\textit{i.e.,} source domain) of an algorithm and the test domain encountered in a practical application (\textit{i.e.,} target domain). Although recent UDA studies~\cite{na2021fixbi,yue2021transporting,xiao2021dynamic} have shown encouraging results, a large domain shift is still a significant obstacle.

One recent paradigm to address the large domain shift problem is to leverage intermediate domains between the source and target domains instead of direct domain adaptation. Recent studies~\cite{gong2019dlow,chang2019all} inspired by generative adversarial networks~\cite{goodfellow2014generative} (GANs) generate instances of intermediate domains to bridge the source and target domains. 
% To avoid time-consuming generative methods, 
Moreover, \cite{cui2020gradually,bousmalis2016domain} learn domain-invariant representations by borrowing only the concept of adversarial training. Meanwhile, with the development of data augmentation techniques, many approaches have emerged built on data augmentation to construct the intermediate spaces. Recent studies~\cite{na2021fixbi,xu2020adversarial,wu2020dual} have shown promising results by grafting Mixup augmentation~\cite{zhang2017mixup} to the domain adaptation task. These studies use inter-domain mixup to efficiently overcome the domain shift problem by utilizing vicinal instances between the source and target domains. However, none of them consider leveraging the predictions of the vicinal instances in the perspective of self-training~\cite{lee2013pseudo}.

Self-training is the straightforward approach that uses self-predictions of a model to train itself. Semi-supervised learning methods~\cite{lee2013pseudo,pham2021meta} leverage a model's predictions on unlabeled data to obtain additional information used during training as their supervision. In particular, unsupervised domain adaptation methods~\cite{gu2020spherical,shin2020two,saito2017asymmetric} have shown that 
% the self-training (\textit{i.e.,} Pseudo-label~\cite{lee2013pseudo}) 
pseudo-label for the target domain
can play an important role in alleviating the domain shift problem.

% 새로운 안 ----------------------
In this work, we introduce a new \textbf{Co}ntrastive \textbf{Vi}cinal space-based (CoVi) algorithm that leverages vicinal instances from the perspective of self-training~\cite{lee2013pseudo}. In vicinal space, we observe that the source label is generally dominant over the target label before applying domain adaptation. In other words, even if vicinal instances consist of a higher proportion of target instances than source instances (\textit{i.e.,} target-dominant instances), their one-hot predictions are more likely to be source labels (\textit{i.e.,} source-dominant labels). We define this phenomenon as an \textbf{equilibrium collapse of labels} between vicinal instances. We also discover that the entropy of the predictions is maximum at the points where the equilibrium collapse of labels occurs. Hence, we aim to find and address the points where the entropy is maximized between the vicinal instances. Inspired by the minimax strategy~\cite{fan1953minimax}, we present \textit{EMP-Mixup}, which minimizes the entropy for the \textbf{{entropy maximization point (EMP)}}. Our \textit{EMP-Mixup} adaptively adjusts the Mixup ratio according to the combinations of source and target instances through training.

As depicted in Figure~\ref{fig:01}, we further leverage the EMP as a boundary (\textit{i.e.}, EMP-boundary) to divide the vicinal space into source-dominant and target-dominant spaces. Here, the vicinal instances of the source-dominant space have source labels as their predicted top-1 label. Similarly, the vicinal instances of target-dominant space have target labels as their top-1 label. Taking advantage of these properties, we configure two specialized subspaces to reduce inter-domain and intra-domain discrepancy simultaneously.

First, we construct a \textbf{contrastive space} around the EMP-boundary to ensure that the vicinal instances have contrastive views: source-dominant and target-dominant views. Since the contrastive views share the same combination of source and target instances, they should have the same top-2 labels containing the source and target labels. In addition, under our constraints, the two contrastive views have opposite order of the first and second labels in the top-2 labels. 
% (\textit{i.e., contrastive labels}). 
Inspired by consistency training~\cite{sohn2020fixmatch,berthelot2019remixmatch}, we propose to impose consistency on predictions of the two contrastive views. Specifically, we mitigate inter-domain discrepancy by solving a {\textit{``swapped'' prediction problem}} where we predict the top-2 labels of a contrastive view from the other contrastive view. 

Second, we constrain a \textbf{consensus space} outside of the contrastive space to alleviate the categorical confusion within the intra-domain. 
% In contrast to mitigating inter-domain discrepancy in the contrastive space, we focus on resolving categorical uncertainty of the target domain in the consensus space. 
In this space, we generate target-dominant vicinal instances utilizing multiple source instances as a perturbation to a single target instance. Here, the role of the source instances is not to learn classification information of the source domain but to confuse the predictions of the target instances. We can ensure consistent and robust predictions for target instances by enforcing label consensus among the multiple target-dominant vicinal instances to a single target label.

We perform extensive ablation studies for a detailed analysis of the proposed methods. In particular, we achieve comparable performance to the recent state-of-the-art methods in standard unsupervised domain adaptation benchmarks such as Office-31~\cite{saenko2010adapting}, Office-Home~\cite{venkateswara2017deep}, and VisDA-C~\cite{peng2017visda}. Furthermore, we validate the superiority of our instance-wise approach on the PACS~\cite{li2017deeper} dataset for multi-source domain adaptation. Overall, we make the following contributions:

\vspace{-2mm}
\begin{itemize}
    \item This is the first study in UDA to leverage the vicinal space from the perspective of self-training. We shed light on the problem of the equilibrium collapse of labels in the vicinal space and propose a minimax strategy to handle it.
    \item We alleviate inter-domain and intra-domain confusions simultaneously by dividing the vicinal space into contrastive and consensus spaces.
    \item Our method achieves state-of-the-art performance and is further validated through extensive ablation studies.
\end{itemize}

% \vspace{-6mm}
\section{Related Work}
% \vspace{-4mm}
\textbf{Unsupervised domain adaptation.}
One of the representative domain adaptation approaches~\cite{ganin2015unsupervised,xie2018learning} is learning a domain-invariant representation by aligning the global distribution between the source and target domains.
% One of the representative domain adaptation methods~\cite{ganin2015unsupervised, xie2018learning, ganin2016domain} is learning a domain invariant representation or aligning the global distribution between the source and target domains. 
Of particular interest, Xie~\textit{et al.}~\cite{xie2018learning} presented a moving semantic transfer network that aligns labeled source centroids and pseudo-labeled target centroids to learn semantic representations for unlabeled target data. Following~\cite{na2021fixbi,chang2019domain,gu2020spherical}, we adopt this simple but efficient method as our baseline.

Our work is also related to the domain adaptation approaches that consider the inter-domain and intra-domain gap together. Kang~\textit{et al.}~\cite{kang2019contrastive} proposed to minimize the intra-class discrepancy and maximize the inter-class discrepancy to perform class-aware domain alignment. Pan~\textit{et al.}~\cite{pan2020unsupervised} presented a semantic segmentation method that minimizes both inter-domain and intra-domain gaps. Unlike these methods, we introduce a practical approach that uses two specialized spaces to reduce inter-domain and intra-domain discrepancy for each.

\textbf{Mixup augmentation.}
Mixup~\cite{zhang2017mixup} is a data-agnostic and straightforward augmentation using a linear interpolation between two data instances. The Mixup has been applied to various tasks and shown to improve the robustness of neural networks. The recent semi-supervised learning methods~\cite{berthelot2019mixmatch,sohn2020fixmatch,berthelot2019remixmatch} efficiently utilized Mixup to leverage unlabeled data. Meanwhile, several domain adaptation methods~\cite{wu2020dual,xu2020adversarial,na2021fixbi} with Mixup were proposed to alleviate the domain-shift problem successfully. Xu~\textit{et al.}~\cite{xu2020adversarial} and Wu~\textit{et al.}~\cite{wu2020dual} showed promising results using inter-domain Mixup between source and target domains. Recently, Na~\textit{et al.}~\cite{na2021fixbi} achieved a significant performance gain by using two networks trained with two fixed Mixup ratios.
Moreover, the latest studies~\cite{guo2019Mixup,zhu2020automix,mai2021metaMixup} suggested adaptive Mixup techniques instead of using manually designed interpolation policies. For example, Zhu~\textit{et al.}~\cite{zhu2020automix} introduced a more advanced interpolation technique that seeks the Wasserstein barycenter between two instances and proposed an adaptive Mixup.
% Then they generalize to the automatic Mixup by using a learnable discriminator and generator networks. 
Mai~\textit{et al.}~\cite{mai2021metaMixup} introduced a meta-learning-based optimization strategy for dynamically learning the interpolation policy in semi-supervised learning. 
% However, note that unsupervised domain adaptation methods still count on hand-tuned or random interpolation policies.
However, unsupervised domain adaptation methods still count on hand-tuned or random interpolation policies. In this work, we derive the Mixup ratio according to the convex combinations of source and target instances.

% In unsupervised domain adaptation, relying on the random interpolation policy has the following limitations. First, reducing the discrepancy of instance-level between the source and target domains is as important as alleviating the global discrepancy. This is because the degree of domain shift is different from instance to instance, even if they belong to the same domain. However, the traditional Mixup does not consider the characteristics of instance-level at all. Second, extremely target-biased Mixup can negatively affect learning, as noted in Na~\textit{et al.}~\cite{na2021fixbi}, due to the practice of using inaccurate target labels as pseudo-labels. From these perspectives, in this work, we derive the Mixup ratio according to the convex combinations of source and target instances instead of using a random interpolation policy.

\textbf{Consistency training.}
Consistency training is one of the promising components for leveraging unlabeled data, which enforces a model to produce similar predictions of original and perturbed instances. The recent semi-supervised learning methods~\cite{sohn2020fixmatch,berthelot2019remixmatch,berthelot2019mixmatch} utilize unlabeled data by assuming that the model should output similar predictions when fed perturbed versions of the same instance. Berthelot~\textit{et al.}~\cite{berthelot2019mixmatch} applied augmentations several times for each unlabeled instance and averaged them to produce guessed labels. In ReMixMatch~\cite{berthelot2019remixmatch}, Berthelot~\textit{et al.} used the model's prediction for a weakly-augmented instance as the guessed label for multiple strongly-augmented variants of the same instance. Recently, Sohn~\textit{et al.}~\cite{sohn2020fixmatch} encouraged predictions from strongly-augmented instances to match pseudo-labels generated from weakly-augmented instances. Although effective, these methods rely on augmentation techniques such as random augmentation, AutoAugment~\cite{cubuk2018autoaugment}, RandAugment~\cite{cubuk2020randaugment}, and CTAugment~\cite{berthelot2019remixmatch}. By contrast, our method is free from these augmentation techniques and does not require carefully selected combinations of augmentations. We solely leverage mixup augmentation~\cite{zhang2017mixup} to generate the vicinal spaces and achieve the effect of consistency training.

\section{Methodology}

\begin{figure}[t]
\centering
\includegraphics[width=0.6\columnwidth]{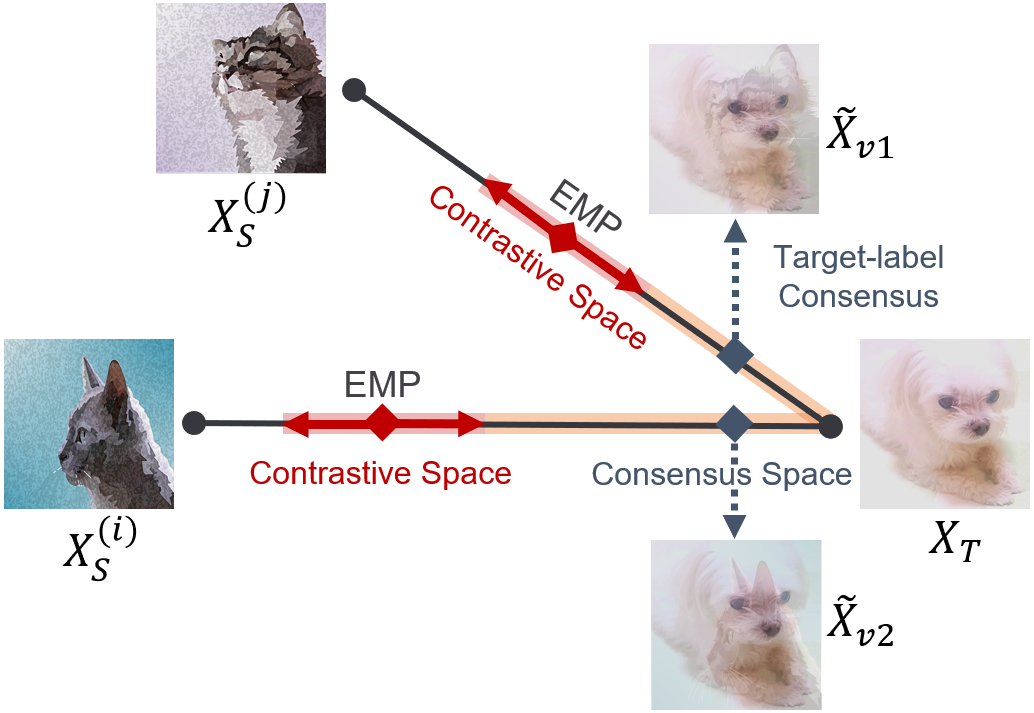}
\vspace{-2mm}
\caption{\textbf{Schematic illustration of CoVi.} The \textit{EMP-Mixup} finds the most confusing point (\textit{i.e.}, EMP) among vicinal instances. 
CoVi then learns through top-\textit{k} contrastive predictions from contrastive views in the contrastive space determined by the EMP. In the consensus space, we achieve a target-label consensus with perturbations of the source instances. 
% Best viewed in color.
}
\label{fig:Overall}
\vspace{-5mm}
\end{figure}

CoVi introduces three techniques to leverage the vicinal space between the source and target domains: i) EMP-Mixup, ii) contrastive views and labels, and iii) a label-consensus. An overall depiction of CoVi is in Figure~\ref{fig:Overall}.
% \vspace{-2mm}
\subsection{Preliminaries}
% \paragraph{Notation.} 
\textbf{Notation.} We denote a mini-batch of $m$-images as $\mathcal{X}$, corresponding labels as $\mathcal{Y}$, and extracted features from $\mathcal{X}$ as $\mathcal{Z}$. Specifically, $\mathcal{{X}_S}\subset\mathbb{R}^{m\times i}$ and $\mathcal{{Y}_S}\subset\{0,1\}^{m\times n}$ denote the mini-batches of source instances and their corresponding one-hot labels, respectively. Here, $n$ denotes the number of classes and $i=c\cdot h\cdot w$, where $c$ denotes the channel size, and $h$ and $w$ denote the height and width of the image instances, respectively. Similarly, the mini-batch of unlabeled target instances is $\mathcal{{X}_T}\subset\mathbb{R}^{m\times i}$. Our model consists of the following subcomponents: an \textit{encoder} $f_\theta$, a \textit{classifier} $h_\theta$, and an \textit{EMP-learner} $g_\phi$.

% \vspace{-3mm}
% \paragraph{Mixup.} 
\noindent\textbf{Mixup.} The \textit{Mixup} augmentation \cite{zhang2017mixup} based on the Vicinal Risk Minimization (VRM) \cite{chapelle2001vicinal} principle exploits virtual instances constructed with the linear interpolation of two instances. These vicinal instances can benefit unsupervised domain adaptation, which has no target domain labels. We define the inter-domain Mixup applied between the source and target domains as follows:

\begin{equation}
\label{eqn:eq1}
\begin{gathered}
    \mathcal{\Tilde{X_\lambda}} = \lambda\cdot\mathcal{{X}_S} +(1-\lambda)\cdot\mathcal{{X}_T} \\
    \mathcal{\Tilde{Y_\lambda}} = \lambda\cdot\mathcal{{Y}_S} +(1-\lambda)\cdot\mathcal{\hat{Y}_T},
\end{gathered}
\end{equation}
where $\mathcal{\hat{Y}_T}$ denotes the pseudo labels of the target instances and $\lambda\in [0,1]$ is the Mixup ratio. Then, the empirical risk for vicinal instances in the inter-domain Mixup is defined as follows:
\begin{equation}
    \mathcal{R}_{\lambda}= \frac{1}{m}\sum_{i=1}^{m}{\mathcal{H}[h(f(\mathcal{\Tilde{X}_\lambda}^{(i)})), \mathcal{\Tilde{Y}_\lambda}^{(i)}]},
\end{equation}
where $\mathcal{H}$ is a standard cross-entropy loss.

\subsection{EMP-Mixup}
In the vicinal space between the source and target domains, we make interesting observations on unsupervised domain adaptation.\\
% \noindent \fbox{\parbox{\textwidth}{\textbf{Observation 1.} \emph{``The labels of the target domain are relatively recessive than the source domain labels.''}}}\\
\noindent\textbf{Observation 1.} \emph{``The labels of the target domain are relatively recessive to the source domain labels.''}

\indent We investigate the dominance of the predicted top-1 labels between the source and target instances in vicinal instances. We find that the label dominance is balanced when the labels of both the source and target domains are provided (\textit{i.e.,} supervised learning). In this case, the top-1 label of the vicinal instance is determined by the instance occupying a relatively larger proportion. However, in the UDA, where the label of the target domain is not given, the balance of label dominance is broken (\textit{i.e.,} equilibrium collapse of labels). Indeed, we discover that source labels frequently represent vicinal instances even with a higher proportion of target instances than source instances.

% \noindent \fbox{\parbox{\textwidth}{\textbf{Observation 2.} 
% \emph{``Depending on the convex combinations of source and target instances, the label dominance is changed.''}}}\\
% \emph{``The label dominance is changed depending on the convex combinations of source and target instances.''}\\
\noindent\textbf{Observation 2.} 
\emph{``Depending on the convex combinations of source and target instances, the label dominance is changed.''}

\indent Next, we observe that the label dominance is altered according to the convex combinations of instances. It implies that an instance-wise approach can be a key to solving the label equilibrium collapse problem. In addition, we discover that the entropy of the prediction is maximum at the point where the label dominance changes because the source and target instances become most confusing at this point (see Figures~\ref{fig:gradcam} and \ref{fig:emp}). 

Based on these observations, we aim to capture and mitigate the most confusing points, which vary with the combination of instances. Inspired by \cite{fan1953minimax,miyato2018virtual,goodfellow2014generative}, we introduce a minimax strategy to break through the worst-case risk~\cite{fan1953minimax} among the vicinal instances between the source and target domains. We minimize the worst risk by finding the \textbf{entropy maximization point (EMP)} among the vicinal instances. In order to estimate the EMPs, we introduce a small network, \textit{EMP-learner}. This network aims to generate Mixup ratios that maximize the entropy of the encoder $f_\theta$ (\textit{e.g., ResNet}) followed by a classifier $h_\theta$. 

Given $\mathcal{{X}_S}$ and $\mathcal{{X}_T}$, we obtain the instance features $\mathcal{{Z}_S}=f_\theta(\mathcal{{X}_S})$ and $\mathcal{{Z}_T}=f_\theta(\mathcal{{X}_T})$ from the encoder $f_\theta$. Then, we pass the concatenated features $\mathcal{{Z}_S}\oplus \mathcal{{Z}_T}$ to the \textit{EMP-learner} $g_\phi$. Then, the \textit{EMP-learner} produces the entropy maximization ratio $\lambda^*$ that maximizes the entropy of the encoder $f_\theta$. Formally, the Mixup ratios for our \textit{EMP-Mixup} are defined as follows: 
\begin{equation}
\begin{gathered}
    \lambda^{*} = \operatorname*{arg\,max}_{\lambda \in [0, 1]}
    \mathcal{H}[h_\theta(f_\theta(\mathcal{\Tilde{X}_\lambda}))],\\
\end{gathered}
\end{equation}
where $\lambda =g_{\phi}(\mathcal{{Z}_S}\oplus \mathcal{{Z}_T})$ and $\mathcal{H}$ is the entropy loss. 

Finally, we design the objective function for \textit{EMP-learner} to \textbf{maximize the entropy} as follows:
\begin{equation}
    \mathcal{R}_\lambda(\phi)= \frac{1}{m}\sum_{i=1}^{m}{
    % f(\mathcal{\Tilde{X}}_{\lambda}^{(i)}) \log(f(\mathcal{\Tilde{X}}_{\lambda}^{(i)}))
    \mathcal{H}[h(f(\mathcal{\Tilde{X}}_{\lambda}^{(i)}))]
    },
\end{equation}
where $\mathcal{H}$ is the entropy loss. Note that we only update the parameter $\phi$ of the \textit{EMP-learner}, not the parameter $\theta$ of the encoder and the classifier.
With the worst-case ratio $\lambda^*$, \textit{EMP-Mixup} \textbf{minimizes the worst-case risk} on vicinal instances as follows:
\begin{equation}
    \mathcal{R}_{\lambda^*}(\theta)= \frac{1}{m}\sum_{i=1}^{m}{\mathcal{H}[h(f(\mathcal{\Tilde{X}}_{\lambda^*}^{(i)})), \mathcal{\Tilde{Y}}_{\lambda^*}^{(i)}]},
\end{equation}
where $\mathcal{H}$ is the standard cross-entropy loss. 

% It is worth noting that
It is noteworthy that our $\lambda^*=[\lambda_1,...,\lambda_m]$ has different optimized ratios according to the combinations of the source and target instances within a mini-batch. Finally, \textit{EMP-Mixup} minimizes the risk of vicinal instances from the viewpoint of the worst-case risk. The overall objective functions are defined as follows:
\begin{equation}
    \mathcal{R}_{emp} = \mathcal{R}_{\lambda^*}(\theta) - \mathcal{R}_{\lambda}(\phi).
\end{equation}

\subsection{Contrastive Views and Labels}

% \noindent \fbox{\parbox{\textwidth}{\textbf{Observation 3.} \emph{``The dominant/recessive labels of the vicinal instances are switched at the EMP.''}}}\\
\textbf{Observation 3.} \emph{``The dominant/recessive labels of the vicinal instances are switched at the EMP.''}\\
\indent Looking back to the previous observations, the label dominance 
% changes according to 
depends on the convex combination of instances, and the point of change is the EMP. In other words, with the EMP as a boundary (\textit{i.e., EMP-boundary}), the dominant/recessive label is switched between the source and target domains. 
It means that vicinal instances around the \textit{EMP-boundary} should have source and target labels as their top-2 labels.
% Hence, vicinal instances around the \textit{EMP-boundary} should have source and target labels as their top-2 labels.

\begin{figure}[t]
\centering
\includegraphics[width=0.7\columnwidth]{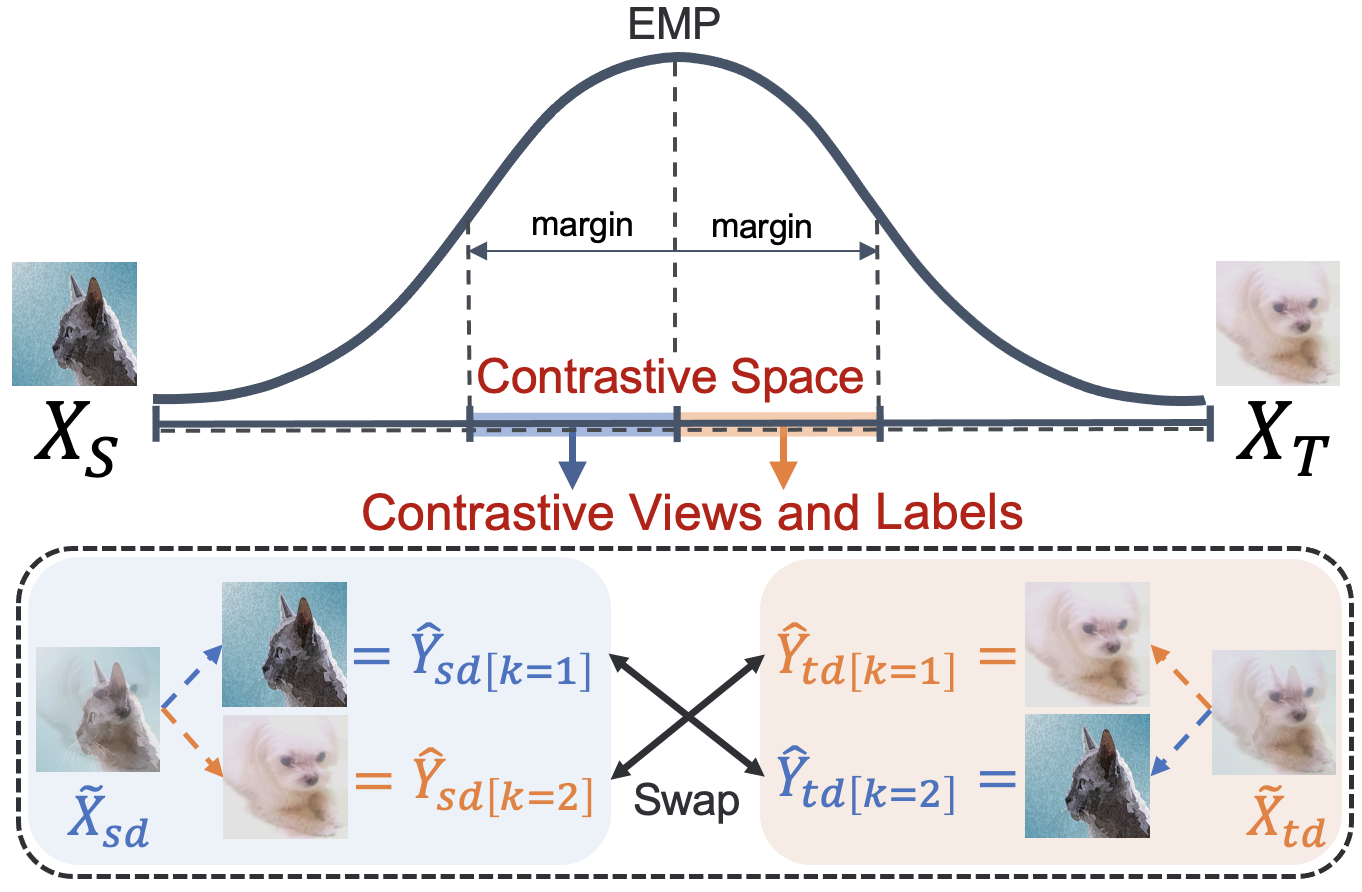}
\vspace{-2mm}
\caption{\textbf{Contrastive views and labels.} \lowercase\expandafter{(\romannumeral1)}~The~\textbf{contrastive views} consist of a source-dominant view $\mathcal{\Tilde{X}}_{sd}$ and a target-dominant view $\mathcal{\Tilde{X}}_{td}$. \lowercase\expandafter{(\romannumeral2)}~The~\textbf{contrastive labels} comprise the source-dominant label and target-recessive label from the top-2 predictions in the contrastive view $\mathcal{\Tilde{X}}_{sd}$ (and vice versa). 
% We enforce consistency between contrastive labels by solving the "swapped" prediction problem.
}
\label{fig:ContrastiveSpace}
\vspace{-4mm}
\end{figure}

These observations and analyses lead us to design the concepts of contrastive views and contrastive labels. Owing to the \textit{EMP-boundary}, we can divide the vicinal space into source-dominant and target-dominant space, as described in Figure~\ref{fig:ContrastiveSpace}. 
% Consequently, the source-dominant instances $\mathcal{\Tilde{X}}_{sd}$ and target-dominant instances $\mathcal{\Tilde{X}}_{td}$ have \textbf{contrastive views} of each other.
Specifically, we constrain the source-dominant and target-dominant spaces of the contrastive space to $\lambda^*-\omega < \lambda_{sd} < \lambda^*$ and $\lambda^* < \lambda_{td} < \lambda^*+\omega$, respectively. Here, $\omega$ is the margin of the ratio from the \textit{EMP-boundary}, which is manually designed. Consequently, the source-dominant instances $\mathcal{\Tilde{X}}_{sd}$ and target-dominant instances $\mathcal{\Tilde{X}}_{td}$ have \textbf{contrastive views} of each other.

From the contrastive views, we focus on the top-2 labels for each prediction because we are only interested in the classes that correspond to the source and target instances, not the other classes. 
Here, we define a set of top-2 one-hot labels within a mini-batch as $\mathcal{\Hat{Y}}_{[k=1]}$ and $\mathcal{\Hat{Y}}_{[k=2]}$. Unlike a general Mixup that uses pure source and target labels (see Eq.\ref{eqn:eq1}), we directly exploit the predicted labels from vicinal instances. In this case, for example, the labels for the instances of the target-dominant space are constructed as follows: 
\begin{equation}
\label{eqn:eq7}
\begin{gathered}
    \mathcal{\Hat{Y}}_{td} = \lambda_{td}\cdot\mathcal{\Hat{Y}}_{td[k=1]} +(1-\lambda_{td})\cdot\mathcal{\Hat{Y}}_{td[k=2]}.
\end{gathered}
\end{equation}

Furthermore, we expand on this and propose a new concept of \textbf{contrastive labels}. We constrain the top-2 labels from the contrastive views as follows:
\begin{itemize}
  \setlength{\itemindent}{0em}
  \item \textit{$\mathcal{\Hat{Y}}_{sd[k=1]}$ from $\mathcal{\Tilde{X}}_{sd}$ and $\mathcal{\Hat{Y}}_{td[k=2]}$ from $\mathcal{\Tilde{X}}_{td}$ must be equal, as the predictions of the source instances.}
  \item \textit{Similarly, $\mathcal{\Hat{Y}}_{sd[k=2]}$ must be equal to $\mathcal{\Hat{Y}}_{td[k=1]}$, as for the predictions of the target instances.}
\end{itemize}

In other words, the dominant label $\mathcal{\Hat{Y}}_{sd[k=1]}$ of $\mathcal{\Tilde{X}}_{sd}$ and the recessive label $\mathcal{\Hat{Y}}_{td[k=2]}$ of $\mathcal{\Tilde{X}}_{td}$ must be the same as the source labels and vice versa. Note that our contrastive constraints are instance-level constraints that must be satisfied between any instances, regardless of the class category. 

Consequently, we swap the top-2 contrastive labels between two contrastive views to learn from the predictions of the other view. By solving a ``swapped'' prediction problem, we enforce consistency to the top-2 contrastive labels obtained from contrastive views of the same source and target instance combinations.

% This method contributes to building a robust model by enforcing consistency between predicted contrastive labels produced for different contrastive views.
According to the constraints, Eq.\ref{eqn:eq7} still holds when we swap the contrastive labels. Finally, the objective for our contrastive loss in target-dominant space is defined as follows:

\begin{equation}
    \mathcal{R}_{td}(\theta)= \frac{1}{m}\sum_{i=1}^{m}{\mathcal{H}[h(f(\mathcal{\Tilde{X}}_{td}^{(i)})), \mathcal{\Hat{Y}}_{td}^{(i)}]},
\end{equation}
\begin{equation}
\begin{gathered}
    \text{where~\; \ }\mathcal{\Hat{Y}}_{\textcolor{OrangeRed}{td}} = \textcolor{OrangeRed}{\lambda_{td}}\cdot\mathcal{\Hat{Y}}_{\textcolor{RoyalBlue}{sd[k=2]}} +(1-\textcolor{OrangeRed}{\lambda_{td}})\cdot\mathcal{\Hat{Y}}_{\textcolor{RoyalBlue}{sd[k=1]}} \nonumber.
\end{gathered}
\end{equation}

\noindent Similarly, we define $\mathcal{R}_{sd}(\theta)$ in the source-dominant space and omit it for clarity. The overall objective functions for contrastive loss are defined as follows:
\begin{equation}
    \mathcal{R}_{ct} =  \mathcal{R}_{td}(\theta) +  \mathcal{R}_{sd}(\theta).
\end{equation}

\subsection{Label Consensus}

Even though the confusion between the source and target instances is crucial in the contrastive space, outside of the contrastive space (\textit{i.e.}, \textbf{consensus space}), we pay more attention to the uncertainty of predictions within the intra-domain than inter-domain instances (see Figure~\ref{fig:area}). Here, we exploit multiple source instances to impose perturbations to target predictions rather than classification information for the source domain. It makes a model more robust to the target predictions by enforcing consistent predictions on the target instances even with the source perturbations.

We construct two randomly shuffled versions of the source instances within a mini-batch. We then apply Mixup with a single target mini-batch to obtain two different perturbed views $v_1$ and $v_2$. 
Here, we set the mixup ratio for the source instances sufficiently small since too strong perturbations can impair the target class semantics.
We compute two softmax probabilities from the perturbed instances $\mathcal{\Tilde{X}}_{v_1}$ and $\mathcal{\Tilde{X}}_{v_2}$ using an encoder, followed by a classifier. Finally, we aggregate the softmax probabilities and yield a one-hot prediction $\mathcal{\Hat{Y}}$. 

% We accomplish \textbf{target-label consensus} on predicting one target-dominant instance perturbed by the two different source instances through the label $\mathcal{\Hat{Y}}$. 
We accomplish \textbf{target-label consensus} by assigning the label $\mathcal{\Hat{Y}}$ to both versions of the perturbed target-dominant instances $\mathcal{\Tilde{X}}_{v_1}$ and $\mathcal{\Tilde{X}}_{v_2}$. 
% Learning two versions of perturbed instances to a single target label allows us to focus on the target domain category rather than the source domain category. 
% Learning two versions of perturbed instances to a single target label allows us to focus on categorical information within the target domain.
% Consistency training from two perturbed instances to a single target label allows us to focus on categorical information for the target domain.
Imposing consistency to differently perturbed instances for a single target label allows us to focus on categorical information for the target domain.
The objective for label consensus on target instances can be defined as follows:
\begin{equation}
\fontsize{9}{10}\selectfont
\begin{gathered}
    \mathcal{R}_{cs}(\theta)= \frac{1}{m}\sum_{i=1}^{m}{[\mathcal{H}(h(f(\mathcal{\Tilde{X}}_{v_1}^{(i)}), \mathcal{\Hat{Y}}^{(i)}))+{\mathcal{H}(h(f(\mathcal{\Tilde{X}}_{v_2}^{(i)}), \mathcal{\Hat{Y}}^{(i)}))]}},
\end{gathered}
\end{equation}
where $\mathcal{H}$ is the cross-entropy loss. 

Note that this approach is also applicable to source-dominant space, but we exclude it from the final loss as it does not significantly affect the performance.

\section{Experiments}
\vspace{-2mm}
We evaluate our method on four popular benchmarks, including Office-31, Office-Home, VisDA-C, and PACS. 
% In particular, the PACS is a multi-source domain adaptation dataset.
Moreover, we validate our method in a multi-source domain adaptation scenario using the PACS dataset.

\noindent\textbf{- Office-31} \cite{saenko2010adapting} contains 31 categories and 4,110 images in three domains: Amazon (A), Webcam (W), and DSLR (D). We verify our methodology in six domain adaptation tasks.

\noindent\textbf{- Office-Home} \cite{venkateswara2017deep} consists of 64 categories and 15,500 images in four domains: Art (Ar), Clipart (Cl), Product (Pr), and Real-World (Rw).

\noindent\textbf{- VisDA-C} \cite{peng2017visda} is a large-scale dataset for synthetic-to-real domain adaptation across 12 categories. It contains 152,397 synthetic images for the source domain and 55,388 real-world images for the target domain.

\noindent\textbf{- PACS} \cite{li2017deeper} is organized into seven categories with 9,991 images in four domains: Photo (P), Art Painting (A), Cartoon (C), and Sketch (S). 
% We evaluate our method on the PACS dataset for multi-source domain adaptation.
\vspace{-2mm}
\subsection{Experimental Setups}
Following the standard UDA protocol~\cite{ganin2016domain,ganin2015unsupervised}, we utilize labeled source data and unlabeled target data. We exploit ResNet-50~\cite{he2016identity,he2016deep} for Office-31 and Office-Home, and ResNet-101 for VisDA-C. For multi-source domain adaptation, we use ResNet-18 in the PACS dataset. We use stochastic gradient descent (SGD) with a momentum of 0.9 in all experiments and follow the same learning rate schedule as in \cite{ganin2015unsupervised}. For the contrastive loss and label consensus loss, we follow the confidence masking policy of \cite{na2021fixbi} that adaptively changes according to the sample mean and standard deviation across all mini-batches. Meanwhile, we design the \textit{EMP-learner} by using four convolutional layers, regardless of the dataset. More detailed information is provided in the supplementary materials.
\vspace{-2mm}
\subsection{Comparison with the State-of-the-Art Methods}
% We validate our method compared with the state-of-the-art methods on three public benchmarks. We report the accuracy of Office-31, Office-Home, and VisDA-C.
We validate our method compared with the state-of-the-art methods on three public benchmarks, including Office-31, Office-Home, and VisDA-C.

\begin{table*}[t]
\begin{center}
% \vspace{-5mm}
\resizebox{\textwidth}{!}{%
\begin{tabular}{ccccccc|c}
% \hline
\specialrule{.1em}{.05em}{.05em} 
Method       & A$\rightarrow$W & D$\rightarrow$W & W$\rightarrow$D & A$\rightarrow$D & D$\rightarrow$A & W$\rightarrow$A                    & Avg.          \\ 
\hline
% ResNet-50~\cite{he2016deep} & 68.4±0.2          & 96.7±0.1       & 99.3±0.1           & 68.9±0.2          & 62.5±0.3 & \multicolumn{1}{c|}{60.7±0.3} & 76.1 \\
% DANN (ICML'15)~\cite{ganin2015unsupervised}   & 82.0±0.4          & 96.9±0.2       & 99.1±0.1           & 79.7±0.4          & 68.2±0.4 & \multicolumn{1}{c|}{67.4±0.5} & 82.2 \\
MSTN* (Baseline)~\cite{xie2018learning}      & 91.3              & 98.9           & \textbf{100.0}     & 90.4              & 72.7     & \multicolumn{1}{c|}{65.6}     & 86.5 \\
DWL (CVPR'21)~\cite{xiao2021dynamic}      & 89.2              & 99.2           & \textbf{100.0}     & 91.2              & 73.1     & \multicolumn{1}{c|}{69.8}     & 87.1 \\
DMRL (ECCV'20)~\cite{wu2020dual}      & 90.8±0.3          & 99.0±0.2       & \textbf{100.0±0.0} & 93.4±0.5          & 73.0±0.3 & \multicolumn{1}{c|}{71.2±0.3} & 87.9 \\
ILA-DA (CVPR'21)~\cite{sharma2021instance}      & 95.72              & 99.25           & \textbf{100.0}     & 93.37              & 72.10     & \multicolumn{1}{c|}{75.40}     & 89.3 \\
MCC (ECCV'20)~\cite{jin2020minimum}      & 95.5±0.2              & 98.6±0.1           & \textbf{100.0±0.0}     & 94.4±0.3              & 72.9±0.2     & \multicolumn{1}{c|}{74.9±0.3}     & 89.4 \\
GSDA (CVPR'20)~\cite{hu2020unsupervised}      & 95.7              & 99.1           & \textbf{100}       & 94.8              & 73.5     & \multicolumn{1}{c|}{74.9}     & 89.7 \\
SRDC (CVPR'20)~\cite{tang2020unsupervised}      & 95.7±0.2          & 99.2±0.1       & \textbf{100.0±0.0}     & 95.8±0.2          & 76.7±0.3 & \multicolumn{1}{c|}{77.1±0.1} & 90.8 \\
RSDA (CVPR'20)~\cite{gu2020spherical} & \underline{96.1±0.2} & \underline{99.3±0.2}       & \textbf{100.0±0.0}       & \underline{95.8±0.3}          & 77.4±0.8 & \multicolumn{1}{c|}{\underline{78.9±0.3}} & 91.1\\
FixBi (CVPR'21)~\cite{na2021fixbi} & \underline{96.1±0.2}            & \underline{99.3±0.2}            & \textbf{100.0±0.0}  & 95.0±0.4            & \textbf{78.7±0.5}   & \multicolumn{1}{c|}{{\textbf{79.4±0.3}}} & 91.4 \\
% \hline
% \rowcolor{lightgray!40}
\rowcolor{myblue}
CoVi (Ours) &\textbf{97.6±0.2}            & \textbf{99.3±0.1}            & \textbf{100.0±0.0}  & \textbf{98.0±0.3}            & \underline{77.5±0.3}   & \multicolumn{1}{c|}{{78.4±0.3}} & \textbf{91.8} \\ 
\specialrule{.1em}{.05em}{.05em} 
\end{tabular}
}
\end{center}
\vspace{-3mm}
\caption{\label{tab:Office-31}Accuracy (\%) on Office-31 for unsupervised domain adaptation (ResNet-50). The best accuracy is indicated in bold, and the second-best accuracy is underlined. * Reproduced by \cite{chang2019domain}.}
\vspace{-7mm}
\end{table*}

\begin{table*}[t]
% \vspace{-2mm}
\resizebox{\textwidth}{!}{%
\begin{tabular}{ccccccccccccc|c}
% \hline
\specialrule{.1em}{.05em}{.05em} 
Method &
  Ar$\rightarrow$Cl &
  Ar$\rightarrow$Pr &
  Ar$\rightarrow$Rw &
  Cl$\rightarrow$Ar &
  Cl$\rightarrow$Pr &
  Cl$\rightarrow$Rw &
  Pr$\rightarrow$Ar &
  Pr$\rightarrow$Cl &
  Pr$\rightarrow$Rw &
  Rw$\rightarrow$Ar &
  Rw$\rightarrow$Cl &
  Rw$\rightarrow$Pr &
  Avg. \\ 
  \hline
% ResNet-50~\cite{ResNet1} & 34.9 & 50            & 58            & 37.4          & 41.9 & 46.2 & 38.5          & 31.2 & 60.4        & 53.9 & 41.2 & 59.9 & 46.1 \\
MSTN* (Baseline)~\cite{xie2018learning}      & 49.8 & 70.3          & 76.3 & 60.4          & 68.5 & 69.6 & 61.4          & 48.9 & 75.7        & 70.9 & 55   & 81.1 & 65.7 \\
AADA (ECCV'20)~\cite{yang2020mind}      & 54.0 & 71.3          & 77.5 & 60.8          & 70.8 & 71.2 & 59.1            & 51.8 & 76.9          & 71.0 & 57.4 & 81.8 & 67.0 \\
ETD (CVPR'20)~\cite{li2020enhanced}      & 51.3 & 71.9          & \textbf{85.7} & 57.6          & 69.2 & 73.7 & 57.8            & 51.2 & 79.3          & 70.2 & 57.5 & 82.1 & 67.3 \\
GSDA (CVPR'20)~\cite{hu2020unsupervised}      & \textbf{61.3} & 76.1          & 79.4 & 65.4          & 73.3 & 74.3 & 65            & 53.2 & 80          & 72.2 & 60.6 & 83.1 & 70.3 \\
GVB-GD (CVPR'20)~\cite{cui2020gradually}    & 57   & 74.7          & 79.8 & 64.6          & 74.1 & 74.6 & 65.2          & 55.1 & 81          & 74.6 & 59.7 & 84.3 & 70.4 \\
TCM (ICCV'21)~\cite{yue2021transporting}      & 58.6 & 74.4          & 79.6 & 64.5          & 74.0 & 75.1 & 64.6            & 56.2 & 80.9          & 74.6 & 60.7 & 84.7 & 70.7 \\
RSDA (CVPR'20)~\cite{gu2020spherical} & 53.2 & \underline{77.7}          & \underline{81.3} & 66.4          & 74   & 76.5 & \underline{67.9}          & 53   & \underline{82} & 75.8 & 57.8 & 85.4 & 70.9 \\
SRDC (CVPR'20)~\cite{tang2020unsupervised}      & 52.3 & 76.3          & 81            & \textbf{69.5} & 76.2 & \underline{78}   & \textbf{68.7} & 53.8 & 81.7        & 76.3 & 57.1 & 85   & 71.3 \\
MetaAlign (CVPR'21)~\cite{wei2021metaalign}      & \underline{59.3} & 76.0          & 80.2 & 65.7          & 74.7 & 75.1 & 65.7            & 56.5 & 81.6          & 74.1 & 61.1 & 85.2 & 71.3 \\
FixBi (CVPR'21)~\cite{na2021fixbi} &
  58.1 &  77.3 &  80.4 &  67.7 &  \underline{79.5} &  \textbf{78.1}&
  65.8 &  \underline{57.9} &  81.7 &  \underline{76.4} &  \underline{62.9} &  \textbf{86.7} &
  72.7 \\

% \rowcolor{lightgray!40}
\rowcolor{myblue}
CoVi (Ours) &
58.5 &  \textbf{78.1} &  80.0 &  \underline{68.1} &  \textbf{80.0} &  77.0&
66.4 &  \textbf{60.2} &  \textbf{82.1} &  \textbf{76.6} &  \textbf{63.6} &  \underline{86.5} &
\textbf{73.1} \\ 
\specialrule{.1em}{.05em}{.05em} 

\end{tabular}%
}
\vspace{1mm}
\caption{\label{tab:Office-Home}Accuracy (\%) on Office-Home for unsupervised domain adaptation (ResNet-50). The best accuracy is indicated in bold, and the second-best accuracy is underlined. * Reproduced by \cite{gu2020spherical}.}
\vspace{-4mm}
\end{table*}

\begin{table*}[t]
% \vspace{-2mm}
\resizebox{\textwidth}{!}{%
\begin{tabular}{ccccccccccccc|c}
\specialrule{.1em}{.05em}{.05em} 
Method     & aero & bicycle       & bus  & car  & horse & knife & motor & person & plant & skate & train & truck & Avg.  \\ 
\hline
MSTN* (Baseline)~\cite{xie2018learning}       & 89.3 & 49.5          & 74.3 & 67.6 & 90.1  & 16.6  & \textbf{93.6}  & 70.1   & 86.5  & 40.4  & 83.2  & 18.5  & 65.0 \\
% JAN~\cite{Long2017}        & 75.7 & 18.7          & 82.3 & 86.3 & 70.2  & 56.9  & 80.5  & 53.8   & 92.5  & 32.2  & 84.5  & \underline{54.5}  & 65.7 \\
% ADR~\cite{ADR}        & 87.8 & 79.5          & 83.7 & 65.3 & 92.3  & 61.8  & 88.9  & 73.2   & 87.8  & 60    & 85.5  & 32.3  & 74.8 \\
% DM-ADA~\cite{Minghao2020}     & -    & -             & -    & -    & -     & -     & -     & -      & -     & -     & -     & -     & 75.6 \\
DMRL (ECCV'20)~\cite{wu2020dual}       & -    & -             & -    & -    & -     & -     & -     & -      & -     & -     & -     & -     & 75.5 \\
TCM (ICCV'21)~\cite{yue2021transporting}       & -    & -             & -    & -    & -     & -     & -     & -      & -     & -     & -     & -     & 75.8 \\
DWL (CVPR'21)~\cite{xiao2021dynamic}        & 90.7    & 80.2             & 86.1    & 67.6    & 92.4     & 81.5     & 86.8     & 78.1      & 90.6     & 57.1     & 85.6     & 28.7     & 77.1 \\
CGDM (ICCV'21)~\cite{du2021cross}       & 93.4    & 82.7             & 73.2    & 68.4    & 92.9     & 94.5     & 88.7     & 82.1      & 93.4     & 82.5     & 86.8     & \underline{49.2}     & 82.3 \\
STAR (CVPR'20)~\cite{lu2020stochastic}       & 95   & 84            & 84.6 & 73   & 91.6  & 91.8  & 85.9  & 78.4   & 94.4  & 84.7  & 87    & 42.2  & 82.7 \\
CAN (CVPR'19)~\cite{kang2019contrastive} &
  \textbf{97} &  \underline{87.2} &  82.5 &  74.3 &  \textbf{97.8} &  \textbf{96.2} &  90.8 &  80.7 &  \underline{96.6} &  \textbf{96.3} &
  87.5 &  \textbf{59.9} &  \underline{87.2} \\

FixBi (CVPR'21)~\cite{na2021fixbi} &
  96.1 &  \textbf{87.8} &  \textbf{90.5} &  \textbf{90.3} &
  \underline{96.8} &  \underline{95.3} &  \underline{92.8} &  \textbf{88.7} &
  \textbf{97.2} &  \underline{94.2} &  \underline{90.9} &  25.7 &
  \underline{87.2} \\
\rowcolor{myblue}
CoVi (Ours) &
  \underline{96.8} &
  85.6 &
  \underline{88.9} &
  \underline{88.6} &
  \textbf{97.8} &
  93.4 &
  91.9 &
  \underline{87.6} &
  96.0 &
  93.8 &
  \textbf{93.6} &
  48.1 &
  \textbf{88.5} \\
\specialrule{.1em}{.05em}{.05em} 

\end{tabular}%
}
\vspace{1mm}
\caption{\label{tab:VisDA}Accuracy (\%) on VisDA-C for unsupervised domain adaptation (ResNet-101). The best accuracy is indicated in bold, and the second-best accuracy is underlined. * Reproduced by \cite{chang2019domain}.}
\vspace{-7mm}
\end{table*}

\textbf{Office-31.}
In Table~\ref{tab:Office-31}, we show the comparative performance on ResNet-50. 
% We achieve 91.8\% accuracy, which outperforms other state-of-the-art methods. 
We achieve an accuracy of 91.8\%, which is 5.3\% higher than the baseline MSTN~\cite{xie2018learning}, surpassing other state-of-the-art methods.
Our method performs best in four out of six situations, \textit{e.g.,} A$\rightarrow$W, D$\rightarrow$W, W$\rightarrow$D, and A$\rightarrow$D tasks. In particular, in A$\rightarrow$W and A$\rightarrow$D, although the performance improvement of the recent methods has stagnated, our method achieves a significant performance gain.
% the performance improvement of recent methods is stagnant but our method shows a significant performance gain. 
% Compared to the DMRL~\cite{wu2020dual} and FixBi~\cite{na2021fixbi}, which utilize Mixup~\cite{zhang2017mixup}, we also obtain better performance.
We also attain better performance than the Mixup-based methods, \textit{i.e.}, DMRL~\cite{wu2020dual} and FixBi~\cite{na2021fixbi}.
% Compared to the Mixup-based methods i.e., DMRL~\cite{wu2020dual} and FixBi~\cite{na2021fixbi}, we also obtain better performance.

\textbf{Office-Home.} 
Table~\ref{tab:Office-Home} demonstrates the comparison results on the Office-Home dataset based on ResNet-50. Our method achieves the highest accuracy in half of the tasks and is the first to break the 73\% barrier. In particular, we attain over 10\% higher performance from the baseline in Cl$\rightarrow$Pr and Pr$\rightarrow$Cl. In addition, our method outperforms MetaAlign~\cite{wei2021metaalign}, which uses meta-learning schemes, and FixBi~\cite{na2021fixbi}, which operates two backbone networks (\textit{i.e.,} ResNet).
% ensembles the two networks' outputs.

\textbf{VisDA-C.}
In Table~\ref{tab:VisDA}, we validate our method on a large visual domain adaptation challenge dataset with ResNet-101. Our method outperforms the state-of-the-art methods with an accuracy of 88.5\%. Compared to the baseline MSTN~\cite{xie2018learning}, our method achieves a performance improvement of over 23\%. In addition, our method shows better performance than the mixup-based DMRL~\cite{wu2020dual} and FixBi~\cite{na2021fixbi}.
We could not achieve the best accuracy across all categories due to the poor accuracy of the baseline (65.0\%), yet the overall score supports the effectiveness of our method.
\vspace{-2mm}

\subsection{Ablation Studies and Discussions} \label{sec:ablation}

\begin{figure*}[t]
\centering
\includegraphics[width=\linewidth]{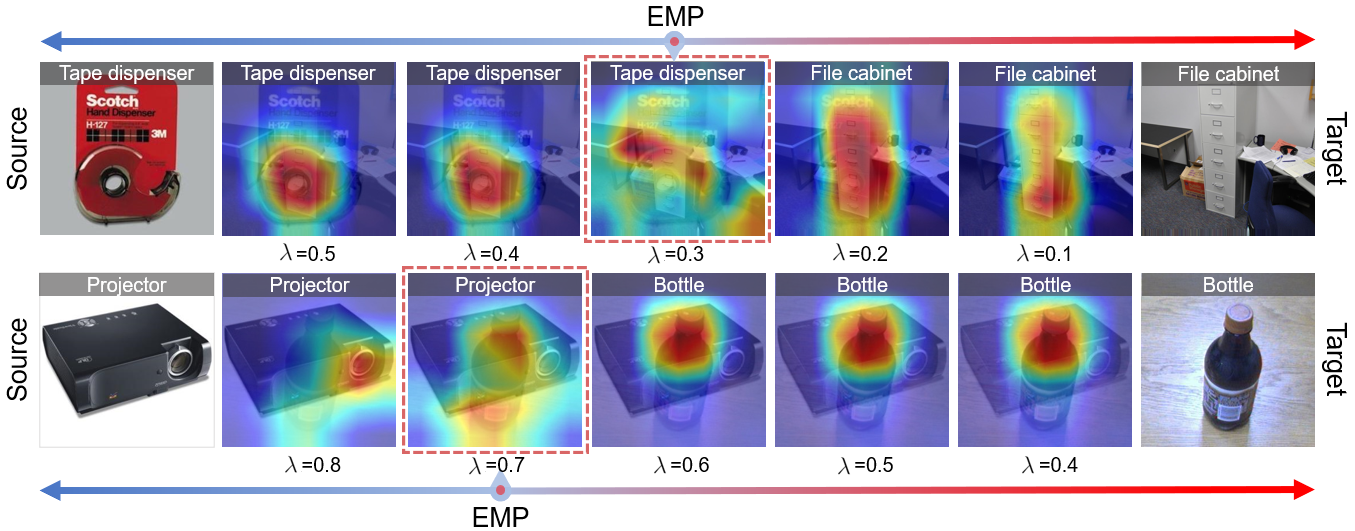}
\vspace{-6mm}
\caption{\textbf{Grad-CAM visualization.} Our key observations in the vicinal space are as follows: \lowercase\expandafter{(\romannumeral1)} \textit{EMPs} vary depending on the convex combination of instances. \lowercase\expandafter{(\romannumeral2)} The \textit{top-1} prediction is switched between the source and target labels (\textit{e.g.,} Tape dispenser $\leftrightarrow$ File cabinet) around the EMP. \lowercase\expandafter{(\romannumeral3)} Grad-CAM highlights the same category as our \textit{top-1} prediction as the most class-discriminative region.
}
\label{fig:gradcam}
\vspace{-4mm}
\end{figure*}

\begin{figure}[t]
% \vspace{-4mm}
% \centering
\begin{minipage}[t]{0.49\linewidth}
% \vspace{0.1mm}
\vspace{0pt}
\includegraphics[width=\columnwidth]{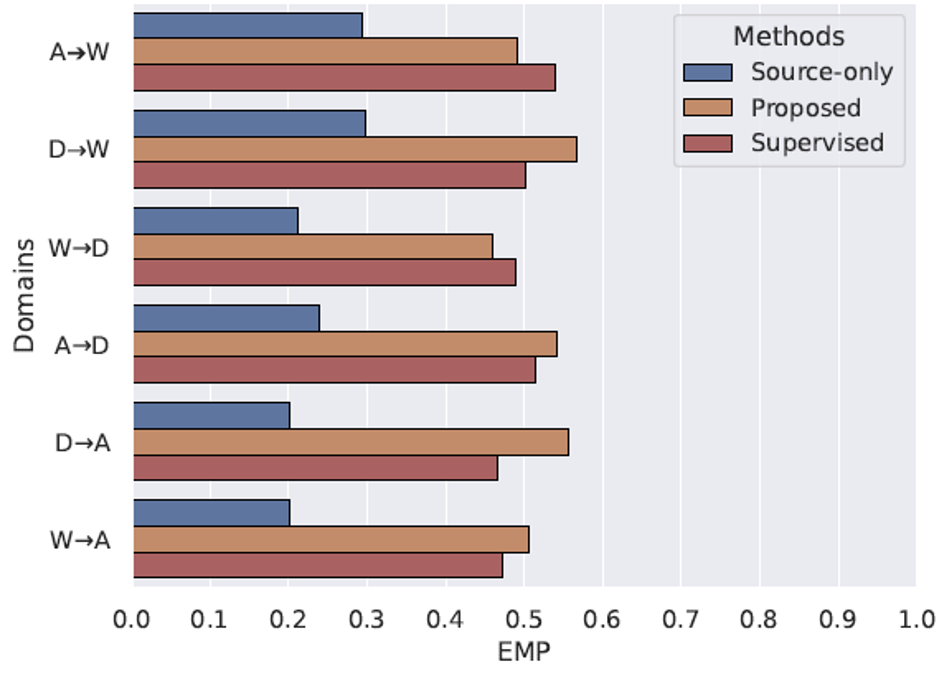}
\vspace{-6.5mm}
\caption{\textbf{Equilibrium collapse of labels.} We compare the change of entropy maximization point according to the methods. Before adaptation, the source domain is dominant over the target domain. Contrarily, applying our method equilibrates around 50\%, similar to supervised learning.}
\label{fig:emp}
\end{minipage}%
\hfill
% \vspace{-3mm}
\begin{minipage}[t]{0.49\linewidth}
\vspace{0pt}
\includegraphics[width=\columnwidth]{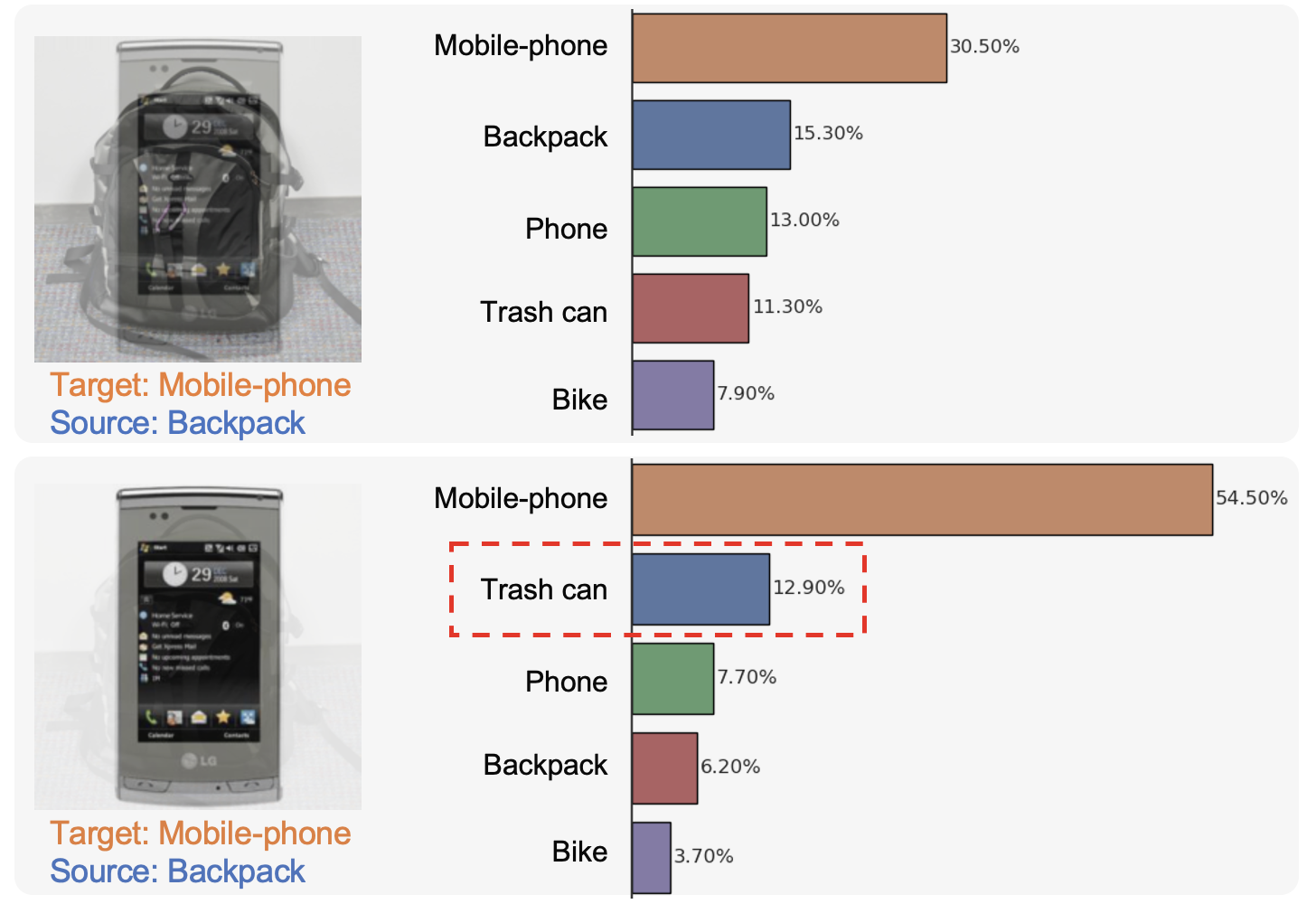}
\vspace{-5mm}
\caption{\textbf{Predictions in the contrastive space \textit{vs.} consensus space.} \textbf{Top:} The first and second predicted labels consist of source and target labels in the contrastive space. \textbf{Bottom:} In the consensus space, the second predicted label is not the source label but a label of another category.}
\label{fig:area}
\end{minipage}%
% \label{fig:area}
% \vspace{-4mm}
\end{figure}

\textbf{Analysis of EMP.} We provide visual examples of the predictions of vicinal instances using Grad-CAM~\cite{selvaraju2017grad} in Figure~\ref{fig:gradcam}. Grad-CAM highlights class-discriminative region in an instance; hence, we can identify the most dominant label in each vicinal instance. Now we demonstrate our crucial observations based on the EMP. First, we observe that the EMP is formed differently depending on the convex combinations of the source and target instances. Second, the dominant labels are switched between the source and target labels at the EMP. Lastly, because the EMP is the highest entropy point, Grad-CAM fails to highlight one specific category at this point adequately. We claim that this outcome is due to the uncertainty arising from the confusion between the source and target instances. Furthermore, we discover that the source and target classes are highlighted in instances on both sides of the EMP.  

\textbf{Equilibrium collapse.}
In Figure~\ref{fig:emp}, we analyze the dominance of labels between the source and target domains. Before adaptation (\textit{i.e.}, source-only), the equilibrium of the labels is broken by the dominant-source and recessive-target domains. In this case, even if the proportion of the target instances in the mixed-up instance is more than half (\textit{i.e.}, target-dominant instance), the top-1 predicted label is determined by the source label (\textit{i.e.}, source-dominant label). In other words, the EMP is formed where it is biased towards the source domain. By contrast, after applying our method, we achieve equilibrium at around 50\%, which is similar to the results of the supervised learning method. Consequently, our method alleviates the equilibrium collapse so that the target-dominant instance is properly predicted as a target label rather than a source label.
% It means that the equilibrium collapse of the label is alleviated so that the target-dominant instance is properly predicted as a target label rather than a source label.
% In other words, the target-dominant instance is predicted as a target label rather than a source label, and the equilibrium collapse of labels is alleviated. 

\textbf{Analysis of the vicinal space.}
Our method leverages the vicinal spaces by dividing them into a contrastive space and a consensus space. In Figure~\ref{fig:area}, we observe that the \textit{top-5} predictions of the two spaces have different characteristics. In the contrastive space, the top-2 predictions consist of the target label (\textit{i.e.}, mobile phone) and source label (\textit{i.e.}, backpack). In other words, the uncertainty between inter-domain categories is the most critical factor in the predictions. By contrast, in the top-2 predictions of the consensus space, the second label is not the source label but another category (\textit{i.e.}, trash can) in the target domain that looks similar to the target label (\textit{i.e.}, mobile phone). Hence, mitigating the intra-domain confusion of the target domain in the consensus space can be another starting point to improve performance further.

\begin{table*}[t]
\vspace{-2mm}
\begin{center}
\resizebox{0.9\textwidth}{!}{%
\begin{tabular}{c|c|c|c||c|c|c|c|c|c|c c}
\specialrule{.1em}{.05em}{.05em} 
\textit{Baseline} & $\mathcal{R}_{emp}$ & $\mathcal{R}_{ct}$ & $\mathcal{R}_{cs}$ & A$\rightarrow$W & D$\rightarrow$W & W$\rightarrow$D & A$\rightarrow$D & D$\rightarrow$A &W$\rightarrow$A & Avg.\\
\hline
\checkmark& &	 &	 &	 91.3&	98.9&	100.0&	90.4&	72.7&	65.6& 86.5\\
\checkmark&	\checkmark&	 & 	&95.9&	99.1&	100.0&	95.6&	76.3&	75.4& 90.4\\
\checkmark& \checkmark&	\checkmark&	 	& 97.1&	99.2&	100.0&	97.2&	76.4&	76.4& 91.1\\
\rowcolor{myblue}
\checkmark& \checkmark&	\checkmark& \checkmark& \textbf{97.6}&	\textbf{99.3}&	\textbf{100.0}&	\textbf{98.0}&	\textbf{77.5}& \textbf{78.4}&	\textbf{91.8}\\
\specialrule{.1em}{.05em}{.05em} 
\end{tabular}
}
\end{center}
\vspace{-2mm}
\caption{\label{tab:Office-31_Ablation}Ablation results (\%) of investigating the effects of our components on Office-31.}
\vspace{-9mm}
\end{table*}

\textbf{Effect of the components.}
We conduct ablation studies to investigate the effectiveness for each component of our method in Table~\ref{tab:Office-31_Ablation}. We observe that our \textit{EMP-Mixup} improves the accuracy by an average of 3.9\% compared to the baseline~\cite{xie2018learning}. In addition, our contrastive loss shows a substantial improvement in the tasks A$\rightarrow$W and A$\rightarrow$D. Meanwhile, in the tasks of D$\rightarrow$A and W$\rightarrow$A, our label-consensus loss significantly impacts the performance gain. Overall, our proposed method improves the baseline by an average of 5.3\%. This experiment verifies that each component contributes positively to performance improvement. 

\textbf{Multi-source domain adaptation.}
To demonstrate the generality of our instance-wise approach, we experiment with a multi-source domain adaptation task, as shown in Table~\ref{tab:PACS}. Our method achieves a performance improvement of over 6\% on the PACS dataset compared to the baseline MSTN~\cite{xie2018learning}. In terms of the average accuracy, our method shows a significant performance improvement compared to the state-of-the-art methods. In particular, our method outperforms three out of four tasks when compared with the recent methods.

\begin{table}[t]
% \vspace{-1mm}
\centering
\resizebox{0.9\columnwidth}{!}{%
\begin{tabular}{ccccc|c}
\specialrule{.1em}{.05em}{.05em} 
Method     & C,S,P$\rightarrow$A & A,S,P$\rightarrow$C       & A,C,P$\rightarrow$S  & A,C,S$\rightarrow$P  &  Avg.  \\ 
\hline
MSTN* (Baseline)~\cite{xie2018learning} &  85.5 & 86.22 & 80.81 & 95.27 & 86.95\\
JiGen (CVPR'19)~\cite{carlucci2019domain} &  86.1 & 87.6 & 73.4 & 98.3 & 86.3\\
Meta-MCD (ECCV'20)~\cite{li2020online} &  87.4 & 86.18 & 78.26 & 97.13 & 87.24\\
CMSS (ECCV'20)~\cite{yang2020curriculum} &  88.6 & 90.4 & 82 & 96.9 & 89.5\\
DSON (ECCV'20)~\cite{seo2020learning} &  86.54 & 88.61 & \underline{86.93} & \textbf{99.42} & 90.38\\
T-SVDNet (ICCV'21)~\cite{li2021t} &  \underline{90.43} & \underline{90.61} & 85.49 & 98.5 & \underline{91.25}\\
\rowcolor{myblue}
CoVi (Ours) & \textbf{93.11} &  \textbf{93.86} &  \textbf{88.06} &  \underline{99.04} & \textbf{93.52}\\
\specialrule{.1em}{.05em}{.05em} 

\end{tabular}%
}
\vspace{1mm}
\caption{\label{tab:PACS}Accuracy (\%) on PACS for multi-source unsupervised domain adaptation (ResNet-18). The best accuracy is indicated in bold, and the second-best accuracy is underlined. * Reproduced by ourselves.}
\vspace{-4mm}
\end{table}

\begin{figure}[t]
    \centering
    % \vspace{-1mm}
    \begin{subfigure}[b]{0.3\linewidth}
        \includegraphics[width=\linewidth]{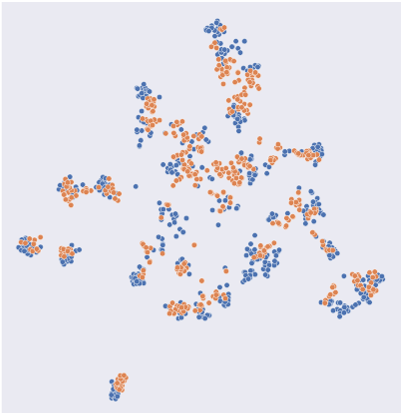}
        \caption{Before Adaptation}
        \label{fig:tsne0}
    \end{subfigure}
    \begin{subfigure}[b]{0.3\linewidth}
        \includegraphics[width=\linewidth]{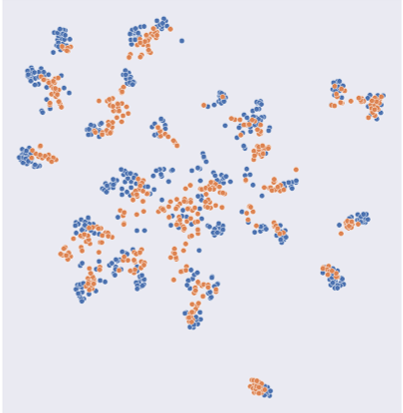}
        \caption{Baseline (MSTN)}
        \label{fig:tsne1}
    \end{subfigure}
    \begin{subfigure}[b]{0.3\linewidth}
        \includegraphics[width=\linewidth]{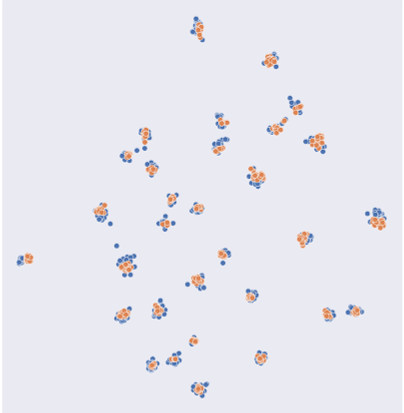}
        \caption{CoVi (Ours)}
        \label{fig:tsne2}
    \end{subfigure}
    \vspace{-2.0mm}
    \caption{\textbf{t-SNE visualization.} Visualization of embedded features on task A$\rightarrow$D. Blue and orange points denote the source and target domains, respectively. 
    % Best viewed in color.
    }
    \label{fig:tsne}
\vspace{-4mm}
\end{figure}

\textbf{Feature visualization.}
We visualize the embedded features on task A$\rightarrow$D of the Office-31 dataset using t-SNE~\cite{van2008visualizing} in Figure~\ref{fig:tsne}. 
Before adaptation, the source embeddings are naturally more cohesive than that of target features because only source supervision is accessible.
% For the baseline MSTN~\cite{xie2018learning} results, we observe that, unlike the source features, the target features do not form cohesive clusters. 
% For the baseline MSTN~\cite{xie2018learning} results, we observe that the target features still fail to form cohesive clusters compared to the source clusters.
After applying the baseline (\textit{i.e.}, MSTN~\cite{xie2018learning}), we observe that the cohesion of the target features is improved but still fails to form tight clusters. By contrast, in our method, the target features construct compact clusters comparable to the source features. These results prove that our method works successfully in the unsupervised domain adaptation task.

% \vspace{-3mm}
\section{Conclusions}
In this study, we investigated the vicinal space between the source and target domains from the perspective of self-training. We raised the problem of the equilibrium collapse of labels and proposed three novel approaches. Our EMP-Mixup efficiently minimized the worst-case risk in the vicinal space. In addition, we reduce inter-domain and intra-domain confusions by dividing the vicinal space into contrastive and consensus space. The competitiveness of our approach suggests that self-predictions in vicinal space can play an important role in solving the UDA problem.\\

\noindent \textbf{Acknowledgement.} 
% This work was partially supported by the MSIT (Ministry of Science, ICT), Korea, under the High-Potential Individuals Global Training Program (2021-0-02130) and Development of High Performance Visual BigData Discovery Platform for Large-Scale Realtime Data Analysis (2014-3-00123) supervised by the IITP (Institute for Information \& Communications Technology Planning \& Evaluation). This work was also partly supported by the BK21 FOUR program of the National Research Foundation of Korea funded by the Ministry of Education (NRF5199991014091).
This work was partially supported by the IITP grant funded by the MSIT, Korea [2014-3-00123, 2021-0-02130] and the BK21 FOUR program (NRF-5199991014091).
\clearpage
% ---- Bibliography ----
%
% BibTeX users should specify bibliography style 'splncs04'.
% References will then be sorted and formatted in the correct style.
%
\bibliographystyle{splncs04}
\bibliography{egbib}
\renewcommand\thesection{\Alph{section}.}
\renewcommand{\thesubsection}{\thesection\arabic{subsection}.}
\renewcommand{\thesubsubsection}{\thesubsection.\arabic{subsubsection}}
\renewcommand{\thefigure}{\Alph{section}.\arabic{subsection}.}
\renewcommand\thetable{\Alph{table}.\arabic{subsection}.}
\counterwithin*{figure}{subsubsection}

\captionsetup[table]{labelsep=space}
\captionsetup[figure]{labelsep=space}

\setcounter{section}{0}
\setcounter{table}{0}
\setcounter{figure}{0}
\begin{titlepage}

% \centerline{\LARGE{\textbf{Supplementary Material}}}

\section{Additional Experimental Results}

\subsection{Effects of our components with a different baseline.}
In the main paper, we provided the effect of our components with baseline, MSTN~\cite{xie2018learning}. We further investigate our method using DANN (Ganin \textit{et al.,} JMLR 2016)~\cite{ganin2015unsupervised} as a baseline, which is one of the simplest methods in unsupervised domain adaptation. As in Table~\ref{tab:dann}, we observed that each component is still effective even with the light baseline DANN. Note that we only obtain the initial weights from the baseline and do not use any losses from the baseline when training our method.
\begin{table}[h]
% \vspace{1mm}
\vspace{-3mm}
\begin{center}
\resizebox{0.99\columnwidth}{!}{%
\begin{tabular}{c|c|c|c||c|c|c|c|c|c|c c}
\specialrule{.1em}{.05em}{.05em} 
\textit{Baseline} & $\mathcal{R}_{emp}$ & $\mathcal{R}_{ct}$ & $\mathcal{R}_{cs}$ & A$\rightarrow$W & D$\rightarrow$W & W$\rightarrow$D & A$\rightarrow$D & D$\rightarrow$A &W$\rightarrow$A & Avg.\\
\hline
\checkmark& &	 &	 &	 82.0&	96.9&	99.1&	79.7&	68.2&	67.4& 82.2\\
\checkmark&	\checkmark&	 & 	& 94.5 &	99.0&	100.0&	94.2&	75.6&	75.2& 89.8\\
\checkmark& \checkmark&	\checkmark&	 	& 95.5&	99.2&	100.0&	94.4&	76.0&	76.3& 90.2\\
% \rowcolor{myblue}
\checkmark& \checkmark&	\checkmark& \checkmark& \textbf{95.6}&	\textbf{99.2}&	\textbf{100.0}&	\textbf{95.8}&	\textbf{76.9}& \textbf{78.3}&	\textbf{91.0}\\
\specialrule{.1em}{.05em}{.05em} 
\end{tabular}
}
\vspace{2mm}
\caption{\label{tab:dann}Ablation results (\%) of investigating the effects of our components with baseline DANN on Office-31.}
% \vspace{-3mm}
\end{center}
\end{table}
\vspace{-12mm}

\subsection{Empirical visualization of vicinal space.}

We computed the entropy of vicinal instances in task A$\rightarrow$W on Office-31 to support the demo Figure~\ref{fig:01} in the main paper. As in Figure~\ref{fig:supp_visualization}\textcolor{red}{a}, we observed that the entropy maximization point (\textit{i.e., }EMP) is biased toward the target domain before adaptation. Here, we define contrastive space within a certain margin from EMP. On the other hand, after applying our method, we observed that the EMP is formed near the center of the source and target domains (see Figure~\ref{fig:supp_visualization}\textcolor{red}{b}).

\begin{figure}[h]
\centering
% \vspace{-1.5mm}
\includegraphics[width=0.95\columnwidth]{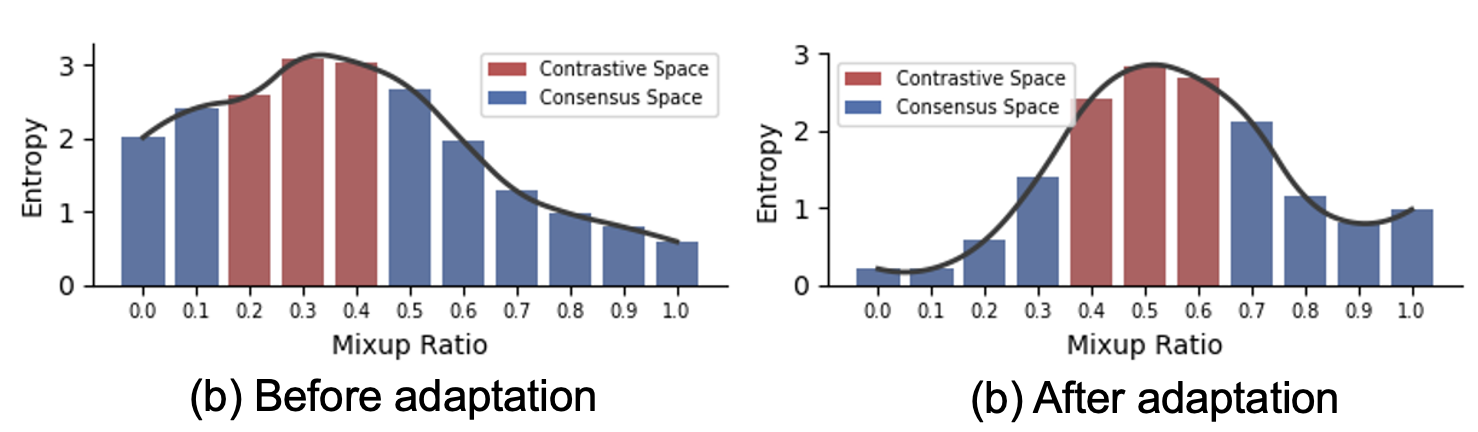}
% \vspace{-4mm}
\caption{Empirical visualization of vicinal space.}
\label{fig:supp_visualization}
\end{figure}

\subsection{The equilibrium collapse of labels in other scenarios.}
As discussed in Section~\ref{sec:ablation}, the equilibrium collapse of labels problem occurs before adaptation by dominant-source and recessive-target vicinal instances. We analyzed whether this problem still exists after applying other UDA methods in Figure~\ref{fig:collapse}\textcolor{red}{a}. In this experiment, we use DANN as a baseline, which has relatively low accuracy (82.2\%). We observe that there is still the problem of the equilibrium collapse of labels in some tasks. 
On the other hand, FixBi (Na \textit{et al.,} CVPR 2021)  (91.4\%) achieved an equilibrium similar to the supervised learning method in all tasks. In addition, we experimented on both single-source and multi-source scenarios in Office-Home and PACS datasets, respectively. As shown in Figure~\ref{fig:collapse}\textcolor{red}{b}, we discovered that the problem of equilibrium collapse of labels occurs in both cases.

\begin{figure}[h]
\centering
\includegraphics[width=0.95\columnwidth]{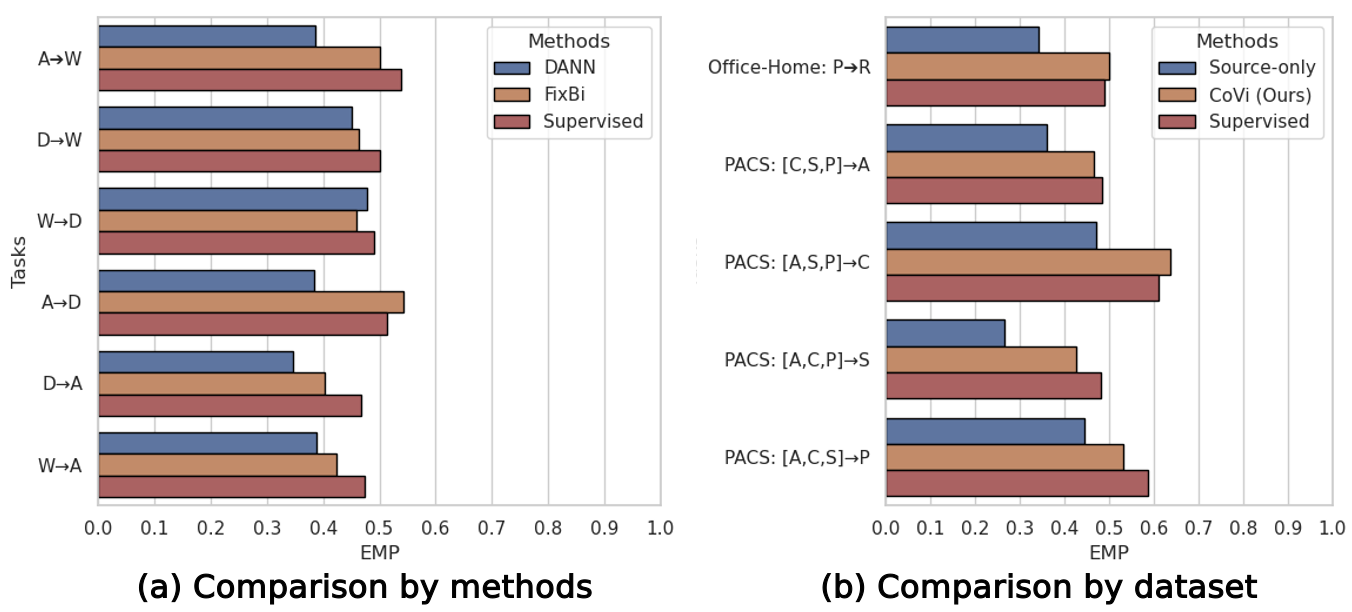}
\caption{Ablation studies on the  `equilibrium collapse of labels'.}
\label{fig:collapse}
\end{figure}
\vspace{-5mm}
\section{Implementation Details}

\subsection{Network Architectures}
We describe the details of network architectures according to the dataset. As introduced in the main paper, our model consists of three subcomponents: an encoder, a classifier, and an EMP-learner. 

\textbf{Encoder.} Following the standard architecture of previous studies on unsupervised domain adaptation~\cite{na2021fixbi,tang2020unsupervised}, we adopt an ImageNet~\cite{krizhevsky2012imagenet}-pretrained ResNet~\cite{he2016identity,he2016deep} for the encoder. 
We use ResNet-50 for Office-31~\cite{saenko2010adapting} and Office-Home~\cite{venkateswara2017deep}, and ResNet-101 for VisDA-C~\cite{peng2017visda} dataset.
% following ~\cite{na2021fixbi}. 
% For Office-31~\cite{saenko2010adapting} and Office-Home~\cite{venkateswara2017deep}, we use ResNet-50. Following ~\cite{na2021fixbi}, we use ResNet-101 for VisDA-C~\cite{peng2017visda} dataset. 
For multi-source domain adaptation, we use ResNet-18 for PACS~\cite{li2017deeper} dataset.

\textbf{EMP-learner.} We introduce a small network to produce entropy maximization points (EMPs) according to the convex combinations of the source and target instances. We design the EMP-learner with four convolutional layers, regardless of the dataset. We construct the EMP-learner with three 3x3 convolutional layers with stride one followed by Batch Normalization~\cite{ioffe2015batch} and ReLU~\cite{nair2010rectified}. For the last layer, instead of the fully connected layer, we adopt 1x1 convolution~\cite{lin2013network}. The output channel of the last 1x1 convolutional layer is 11, yielding a ratio $\lambda \in$ \{0.0, 0.1, ..., 1.0\}.

\textbf{Classifier.} We adopt only one fully connected layer for the classifier. 
% Here, we do not use Batch Normalization~\cite{ioffe2015batch} or ReLU~\cite{nair2010rectified}. 
The input feature size of the fully connected layer is decided by the output feature size of the encoder. The output feature size of the fully connected layer depends on the number of categories in each dataset.

\subsection{Data Configurations}
We implement our algorithm using PyTorch. The code runs with Python 3.7+, PyTorch 1.7.1, and Torchvision 0.8.2. In this section, we provide our training recipes for Office-31, Office-Home, VisDA-C, and PACS dataset in Configs~\ref{Configs:office},\ref{Configs:VisDA-C}, and \ref{Configs:PACS}.

\textbf{Office-31 and Office-Home.} In Configs~\ref{Configs:office}, we describe our default configuration for Office-31 and Office-Home. The default configs for Office-Home are almost identical to Office-31, except for the resize factor of test transform that uses a scaling factor of 256 instead of 224.

\vspace{-4mm}
\begin{Configs}[h]
\caption{\small{PyTorch-style configs for Office-31 and Office-Home.}}
\label{Configs:office}
\begin{MyPython}
train_transforms = torch.nn.Sequential(
    transforms.Resize(256),
    transforms.RandomCrop(224),
    transforms.RandomHorizontalFlip(),
    transforms.ToTensor(),
    transforms.Normalize(
     mean=[0.485, 0.456, 0.406], 
     std=[0.229, 0.224, 0.225]))
test_transforms = torch.nn.Sequential(
    transforms.Resize(224),
    transforms.CenterCrop(224),
    transforms.ToTensor(),
    transforms.Normalize(
     mean=[0.485, 0.456, 0.406], 
     std=[0.229, 0.224, 0.225]))
\end{MyPython}
\end{Configs}
\vspace{-4mm}

\textbf{VisDA-C.} We provide the configurations for VisDA-C in Configs~\ref{Configs:VisDA-C}. 
% We use a training batch size of 128 and an SGD optimizer with a learning rate of 1e-4. 
We use a stochastic gradient descent optimization (SGD) with a training batch size of 128, a momentum of 0.9, and a learning rate of 1e-4.
The end-to-end pipeline is trained for 100 epochs. We use the center crop instead of the random crop for image transformations in the training process. It is worth noting that we do not use the ten-crops ensemble technique used in \cite{xu2019larger,cao2018partial,long2017conditional} during evaluation for a fair comparison. 

% \vspace{-18mm}
\begin{Configs}[h]
\caption{\small{PyTorch-style configs for VisDA-C.}}
\label{Configs:VisDA-C}
\begin{MyPython}
train_transforms = torch.nn.Sequential(
    transforms.Resize(256),
    transforms.CenterCrop(224),
    transforms.RandomHorizontalFlip(),
    transforms.ToTensor(),
    transforms.Normalize(
        mean=[0.485, 0.456, 0.406], 
        std=[0.229, 0.224, 0.225]))
test_transforms = torch.nn.Sequential(
    transforms.Resize(256),
    transforms.CenterCrop(224),
    transforms.ToTensor(),
    transforms.Normalize(
        mean=[0.485, 0.456, 0.406], 
        std=[0.229, 0.224, 0.225]))
\end{MyPython}
\end{Configs}

\indent\textbf{PACS.} Following the previous protocols~\cite{li2021t,seo2020learning} for multi-source domain adaptation, we train on any three of the four domains (i.e., source domains) and then test on the remaining one domain (i.e., target domain). The total epoch is 100, with a batch size of 32 for the PACS dataset. The training details are described in Configs~\ref{Configs:PACS}.

\begin{Configs}[h]
\caption{\small{PyTorch-style configs for PACS.}}
\label{Configs:PACS}
\begin{MyPython}
train_transforms = torch.nn.Sequential(
    transforms.Resize(256),
    transforms.RandomCrop(224),
    transforms.RandomHorizontalFlip(),
    transforms.ToTensor(),
    transforms.Normalize(
        mean=[0.485, 0.456, 0.406], 
        std=[0.229, 0.224, 0.225])
test_transforms = torch.nn.Sequential(
    transforms.Resize(224),
    transforms.CenterCrop(224),
    transforms.ToTensor(),
    transforms.Normalize(
        mean=[0.485, 0.456, 0.406], 
        std=[0.229, 0.224, 0.225]))
        
\end{MyPython}
\end{Configs}
% \end{minipage}

% \vspace{-8mm}
\clearpage
\section{Pseudocode}
\setcounter{algorithm}{0}
\vspace{-3cm}
In Pseudo-code~\ref{alg:EMP-Mixup}, \ref{alg:Contrastive}, and ~\ref{alg:Consensus}, we provide PyTorch-like pseudo-codes for the EMP-Mixup, contrastive loss, and consensus loss, respectively. The entire code has been released at \url{https://github.com/NaJaeMin92/CoVi}.
\vspace{-1.5cm}
\begin{pseudocode}[h]
\flushleft
  \caption{\small{PyTorch-like style pseudocode for EMP-Mixup.}}
%   \caption{\small{Pseudocode of EMP-Mixup in a PyTorch-like style.}}
  \label{alg:EMP-Mixup}
    %   \begin{algorithmic}
    % \NOTATION{In practice, we follow mixup loss to apply EMP-Mixup as mixup\_loss = mixup\_loss(x\_1,y\_1) + mixup\_loss(x\_2,y\_2) (see Eq.1).}
    \NOTATION {x\_s, y\_s: Source image and label}
    \NOTATION {x\_t: Target image}
    \NOTATION {f: An encoder}
    \NOTATION {h: A classifier}
    \NOTATION {g: An EMP-learner}
    \NOTATION {ce\_loss: Cross entropy loss\\}
    % \CODE {for (x\_s, x\_y), (x\_t, $_$) in zip(src\_loader, tgt\_loader)}
    \Comment {\texttt{compute embeddings except for avgpool in f}}
    \CODE {z\_s, z\_t = f(x\_s), f(x\_t)\\}
    \Comment {concat representations along the channel dimension}
    \CODE {z\_c = torch.cat([z\_s, z\_t], dim=1)\\}
    \Comment {Produce entropy maximization points}
    \CODE {emp = torch.argmax(g(z\_c), dim=1) * 0.1\\}
    % \CODE {emp = torch.clamp(emp + m, min=0.0, max=1.0)\\}
    \Comment {construct vicinal instances with EMP}
    \CODE {x\_emp = emp * x\_s + (1 - emp) * x\_t}
    \CODE {z\_emp = h(f(x\_emp))\\}
    \Comment {compute entropy loss}
    \CODE {entropy\_loss = -Entropy(z\_emp)\\}
    \Comment {optimization step}
    \CODE {entropy\_loss.backward()}
    \CODE {update(g.params)\\}
    % \textbf{\texttt{\small{\textcolor{red}{\\Phase 2. Minimize cross-entropy}}}}
    \Comment {compute cross-entropy loss}
    \CODE {y\_t = torch.argmax(h(f(x\_t)), dim=1)}
    % \CODE {ce\_loss = mixup\_loss(z\_emp, y\_s, y\_t, emp)}
    \CODE {mixup\_loss = emp * ce\_loss(z\_emp, y\_s) + (1 - emp) * ce\_loss(z\_emp, y\_t)\\}
    \Comment {optimization step}
    \CODE {mixup\_loss.backward()}
    \CODE {update(f.params)}
    \CODE {update(h.params)}
    %   \end{algorithmic}
\end{pseudocode}
\pagebreak
\begin{pseudocode}[H]
    \caption{\small{PyTorch-like style pseudocode for contrastive loss.}}
%   \caption{\small{Pseudocode of Contrastive loss in a PyTorch-like style.}}
  \label{alg:Contrastive}
    \NOTATION {x\_s, y\_s: Source image and label}
    \NOTATION {x\_t: Target image}
    \NOTATION {f: An encoder}
    \NOTATION {h: A classifier}
    \NOTATION {emp: Entropy maximization point}
    \NOTATION {w: Margin of ratio}
    \NOTATION {alpha: Confidence threshold}
    \NOTATION {space\_sd: Source-dominant space constraint}
    \NOTATION {space\_td: Target-dominant space constraint}
    \NOTATION {ce\_loss: Cross entropy loss}
    \NOTATION {In practice, we replace top2\_sd with y\_hat in swap prediction to take advantage of the higher accuracy top1 label. Also, we replace top2\_td with y\_s because we can access source labels.\\}
    \Comment {\texttt{construct space}}
    \CODE {sd\_ratio, td\_ratio = emp - w, emp + w}
    \CODE {sd\_cont = torch.ge(sd\_ratio, space\_sd)}
    \CODE {td\_cont = torch.le(td\_ratio, space\_td)\\}
    \Comment {compute threshold mask}
    \CODE {z\_t = f(x\_t)}
    \CODE {top1\_prob = torch.topk(F.softmax(z\_t, dim=1), k=1)[0].t().squeeze()}
    \CODE {prob\_mean, prob\_std = top1\_prob.mean(), top1\_prob.std()}
    \CODE {threshold = prob\_mean - alpha * prob\_std}
    \CODE {th\_mask = top1\_prob.ge(threshold)\\}
    \Comment {construct vicinal instances}
    \CODE {mask\_idx = torch.nonzero(th\_mask \& td\_cont \& sd\_cont).squeeze()}
    \CODE {x\_sd = emp[mask\_idx] * x\_s[mask\_idx] + (1 - emp[mask\_idx]) * x\_t[mask\_idx]}
    \CODE {x\_td = emp[mask\_idx] * x\_s[mask\_idx] + (1 - emp[mask\_idx]) * x\_t[mask\_idx]\\}
    \Comment {compute representations}
    \CODE {z\_sd, z\_td = h(f(x\_sd)), h(f(x\_td))\\}
    \Comment {predict top-2 labels}
    \CODE {top1\_sd, top2\_sd = torch.topk(F.softmax(z\_sd, dim=1), k=2)[1].t()}
    \CODE {top1\_td, top2\_td = torch.topk(F.softmax(z\_td, dim=1), k=2)[1].t()\\}
    \Comment {swap predictions and compute contrastive loss}
    \CODE {y\_hat = torch.argmax(h(f(x\_t)), dim=1)}
    \CODE {sd\_loss = sd\_ratio * ce\_loss(z\_sd, top2\_sd) + (1 - sd\_ratio) * ce\_loss(z\_sd, top1\_td)}
    \CODE {td\_loss = td\_ratio * ce\_loss(z\_td, top2\_td) + (1 - td\_ratio) * ce\_loss(z\_td, top1\_sd)\\}
    \CODE {contrastive\_loss = sd\_loss + td\_loss\\}
    \Comment {optimization step}
    \CODE {contrastive\_loss.backward()}
    \CODE {update(f.params)}
    \CODE {update(h.params)}
\end{pseudocode}

\begin{pseudocode}[H]
  \caption{\small{PyTorch-like style pseudocode for consensus loss.}}
%   \caption{\small{Pseudocode of Consensus loss in a PyTorch-like style.}}
  \label{alg:Consensus}
    \NOTATION {x\_s: Source image}
    \NOTATION {x\_t: Target image}
    \NOTATION {f: An encoder}
    \NOTATION {h: A classifier}
    \NOTATION {w: Margin of ratio}
    \NOTATION {beta: Confidence threshold}
    \NOTATION {ce\_loss: Cross entropy loss\\}
    \Comment {\texttt{construct two perturbed versions}}
    \CODE {shuffle\_idx = torch.randperm(batch\_size)}
    \CODE {x\_v1 = lam * x\_s + (1 - lam) * x\_t}
    \CODE {x\_v2 = lam * x\_s[shuffle\_idx] + (1 - lam) * x\_t\\}
    \Comment {construct representations}
    \CODE {z\_v1 = h(f(x\_v1))}
    \CODE {z\_v2 = h(f(x\_v2))\\}
    \Comment {compute threshold mask}
    \CODE {z\_t = f(x\_t)}
    \CODE {top1\_prob = torch.topk(F.softmax(z\_t, dim=1), k=1)[0].t().squeeze()}
    \CODE {prob\_mean, prob\_std = top1\_prob.mean(), top1\_prob.std()}
    \CODE {threshold = prob\_mean - beta * prob\_std}
    \CODE {th\_mask = top1\_prob.ge(threshold)}
    \CODE {mask\_idx = torch.nonzero(th\_mask).squeeze()\\}
    \Comment {Aggregate softmax probabilities}
    \CODE {p = F.softmax(z\_v1, dim=1) + F.softmax(z\_v2, dim=1)\\}
    \Comment {compute consensus loss}
    \CODE {y\_hat = torch.argmax(p, dim=1)}
    \CODE {loss = ce\_loss(z\_v1[mask\_idx], y\_hat[mask\_idx]) + ce\_loss(z\_v2[mask\_idx], y\_hat[mask\_idx])\\}
    \Comment {optimization step}
    \CODE {loss.backward()}
    \CODE {update(f.params)}
    \CODE {update(h.params)}
\end{pseudocode}

\end{titlepage}
\clearpage

\end{document}